\documentclass[letterpaper, 10 pt, journal, twoside]{IEEEtran}
\IEEEoverridecommandlockouts
\usepackage{xcolor,soul,framed} 
\usepackage{amsmath,amsfonts}
\usepackage{algorithmic}
\usepackage{algorithm}
\usepackage{array}
\usepackage[caption=false,font=normalsize,labelfont=sf,textfont=sf]{subfig}
\usepackage{textcomp}
\usepackage{stfloats}
\usepackage{url}
\usepackage{hyperref}
\usepackage{verbatim}
\usepackage{graphicx}
\usepackage{dsfont}
\usepackage{amsmath}
\usepackage{amssymb}

\usepackage{cite}
\usepackage{booktabs}
\DeclareGraphicsExtensions{.pdf,.jpeg,.png}
\definecolor{myGreen}{RGB}{0,150,80}
\let\labelindent\relax 
\usepackage{enumitem}
\setlist[itemize]{topsep=4pt, itemsep=2pt, parsep=0pt}

\begin{document}
\bstctlcite{IEEEexample:BSTcontrol}
    \title{Stability-aware Residual Reinforcement Learning Framework for Robotic Manipulator Disturbance Compensation}
\author{
Jihong Kim\textsuperscript{\normalfont 1},
Joonhyuk Kwon\textsuperscript{\normalfont 1},
Hwa Soo Kim\textsuperscript{\normalfont 2},~\IEEEmembership{Member,~IEEE},
TaeWon Seo\textsuperscript{\normalfont 1},~\IEEEmembership{Senior Member,~IEEE},
and Hyung-Tae Seo\textsuperscript{\normalfont 3},~\IEEEmembership{Member,~IEEE}

\thanks{
This research was supported in part by the Institute of Information \& Communications Technology Planning \& Evaluation (IITP) grant funded by the Korea government (MSIT) (RS-2025-02304968), and in part by the Korea Planning \& Evaluation Institute of Industrial Technology (KEIT) grant funded by the Korea government (MOTIE) (RS-2026-25531069).
}
\thanks{
Jihong Kim, Joonhyuk Kwon, and TaeWon Seo are with the School of Mechanical Engineering, Hanyang University, Seoul 04763, Republic of Korea (e-mail: \{jh108090@hanyang.ac.kr; luciferjoon@hanyang.ac.kr; taewonseo@hanyang.ac.kr). \textit{(Corresponding authors: Hyung-Tae Seo; TaeWon Seo.)}
}
\thanks{
Hwa Soo Kim is with the Department of Mechanical System Design, Kyonggi University, Suwon 16227, South Korea (e-mail: hskim94@kgu.ac.kr).
}
\thanks{
Hyung-Tae Seo is with the School of Electrical Engineering, Kookmin University, Seoul 02707, South Korea (e-mail: htseo@kookmin.ac.kr).
}
}

\maketitle
\begin{abstract}
Although conventional controllers and disturbance observers (DOBs) are the standard for precision tracking in manipulators, they suffer from parameter uncertainty, nonlinear friction, and compound disturbances. This study proposes a \textit{ residual reinforcement learning DOB} framework that pairs an analytical observer with an RL policy. The deterministic baseline operates within a reliable region, whereas the RL policy explicitly targets the residuals that the model cannot capture. To make this compensation disturbance-aware, an estimator network aligns the observation history with a privileged disturbance context, organizing the latent space by disturbance regime and enabling rapid adaptation across disturbance transitions. To guarantee stability, we derived and enforced a state-dependent action bound on the RL policy from an input-to-state stability (ISS) analysis such that the closed loop provably confines the tracking error to a certified envelope for arbitrary policy outputs. Experiments on a 6-DOF manipulator demonstrated consistent improvements in disturbance estimation and tracking, including a 27.8\% tracking-error reduction on real hardware under zero-shot sim-to-real transfer and a 38.0\% reduction under a base-vibration disturbance that was not observed during training.
\end{abstract}


\IEEEpeerreviewmaketitle
\section{Introduction}
\label{sec:introduction}

\IEEEPARstart{T}{he} growing deployment of industrial manipulators in precision assembly, medical assistance, and human-robot collaboration has increased the demand for control techniques that maintain accurate tracking under external disturbances. Manipulators exhibit strongly coupled nonlinear dynamics with six or more degrees of freedom and are subject to payload variations, friction, and unmodeled effects during operation. Ensuring precise end-effector tracking under these conditions is a critical requirement for most practical applications.

A common approach is to combine a model-based controller, such as nonlinear model predictive control (NMPC) \cite{X1}, with a disturbance observer (DOB). The model-based controller theoretically guarantees stability and tracking with respect to the nominal dynamics, whereas the DOB estimates disturbances in real time to enable the controller to compensate for them \cite{X2}. However, by design, conventional DOBs assume a specific class of disturbances and are constructed accordingly. Consequently, owing to model parameter uncertainty, state-dependent nonlinearities such as friction and backlash, payload variations, or compound disturbances spanning multiple frequency bands, DOBs exhibit structural limitations that cannot be totally overcome by any variant.

\begin{figure} 
    \centering
    \includegraphics[width=\linewidth]{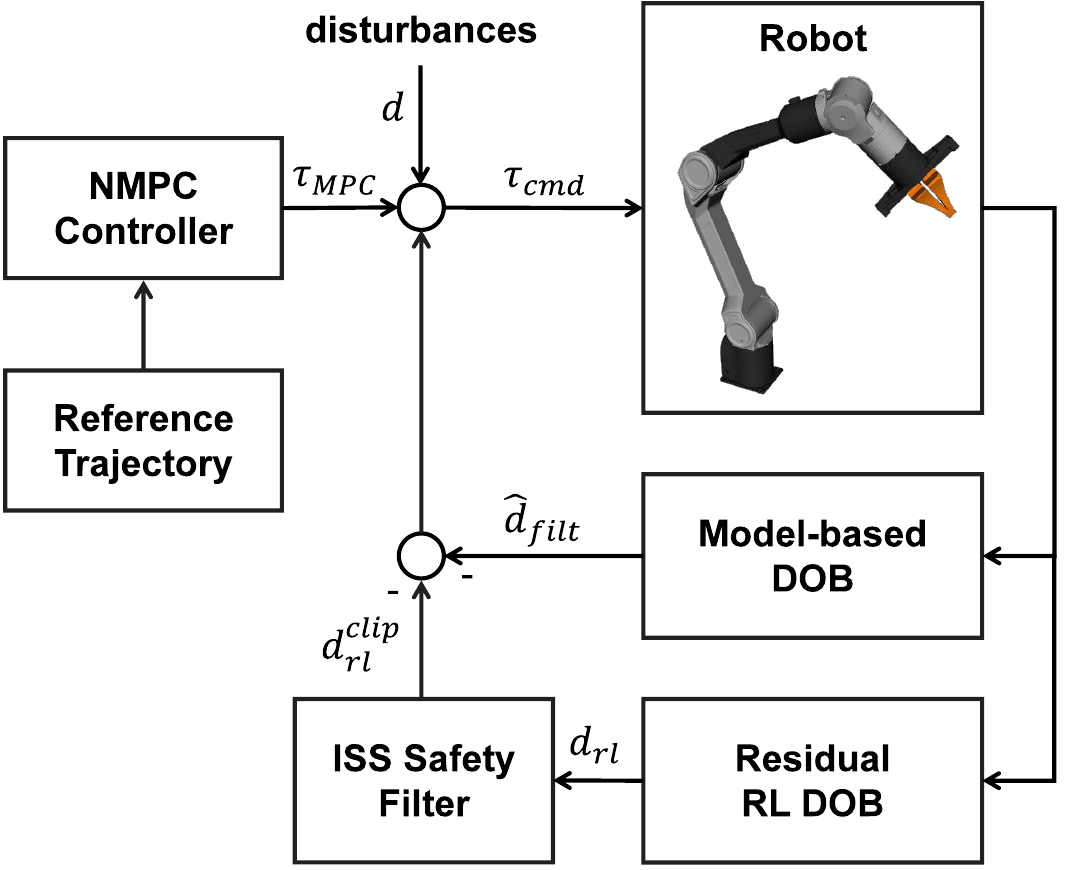} 
    \caption{Overview of the proposed framework. The NMPC and model-based DOB form the nominal loop, and the residual RL policy compensates for the disturbance components that cannot be captured by the DOB.} 
    \label{fig:fig1}
\end{figure}

To address these limitations, reinforcement learning (RL) has been actively investigated for manipulator control. Reinforcement learning inherently adapts to nonlinear and compound disturbances by learning policies directly from the data rather than relying on strict analytical models \cite{X20}. However, their industrial deployment is hindered by two major limitations. First, the sim-to-real gap complicates achieving consistent performance across varying operating regimes when policies are trained in the simulation. Second, real-world training imposes prohibitive demands on data and resources. Fundamentally, these two paradigms are complementary. Model-based controllers provide essential consistency and repeatability through their deterministic design but lack adaptability. By contrast, RL excels in handling unmodeled dynamics but struggles to ensure absolute stability. Although various hybrid approaches have been explored to combine both fields \cite{X4, kamohara2025rl}, designing an integration framework in which the learning policy explicitly preserves the theoretical stability guarantee of the baseline controller remains a critical problem.
 Building on these observations, we proposed a \textit{ residual reinforcement learning DOB} framework. Conventional DOBs estimate the disturbances within a reliable operational range. However, discrepancies inevitably remain between this estimate and the true external disturbance. These gaps include nonlinear friction, model-induced estimation errors, and unestimable components of the compound disturbances. We defined this uncaptured portion as the residual disturbance. 
Instead of replacing the baseline controller entirely, our RL policy was trained explicitly to compensate for this residual. This targeted approach yields faster and more stable convergence than end-to-end RL methods that attempt to learn the entire disturbance dynamics through reward shaping alone. 
Furthermore, it preserves interpretability because the policy output retains a clear physical meaning. To enhance the robustness of this compensation, we integrated an estimator network trained by aligning the observation history with a privileged disturbance context using a prototype-based swapped prediction. This network directly extracts the underlying disturbance characteristics and environmental contexts from the observation history. This estimator accelerates policy training and facilitates rapid adaptation during abrupt environmental transitions by structuring the latent spaces to cluster distinct disturbance types into separate regions. Finally, to ensure that learning-based intervention does not compromise the inherent stability of the system, we derived a state-dependent action bound using input-to-state stability (ISS) analysis. We strictly enforced this bound as a hard limit on the RL policy output. 
The main contributions of this study are summarized as follows:
\begin{itemize}
    \item We propose a residual RL framework in which the policy targets only the disturbance components that model-based DOBs cannot capture, as well as a state-dependent action bound derived from ISS analysis that guarantees a certified error envelope regardless of the policy output.
    \item An estimator network whose latent space is organized by disturbance regime through alignment with a privileged context, enabling the policy to recognize the acting disturbance and adapt rapidly across transitions
    \item Validation through simulation and using a 6-DOF manipulator, which demonstrated improved disturbance estimation, robust tracking under compound disturbances, zero-shot sim-to-real transfer, and generalization to a base-vibration disturbance not encountered during training.
\end{itemize}

\section{Related Work}
\label{sec:related_work}

\subsection{Manipulator Control and Disturbance Observation}
\label{subsec:manipulator_control_dob}

Analytical approaches for precision manipulator control, including computed-torque control (CTC), sliding-mode control, and nonlinear model predictive control (NMPC), have long served as standard baselines by synthesizing inputs directly from nominal dynamics to ensure theoretical stability. NMPC is particularly favored for multi-DOF systems because of its capacity to optimize costs over a finite horizon while managing the state and input constraints \cite{X7}. However, these stability assurances hold strictly under nominal conditions. In physical deployments, manipulators frequently encounter external torques, unmodeled dynamics, and payload variations that violate these assumptions and directly degrade tracking accuracy. Consequently, robust tracking requires active disturbance estimation and control-input compensation.

Disturbance observers (DOBs) are the conventional mechanisms for such compensation. Early frequency-domain DOBs estimate the disturbances by inverting the nominal plant and applying a Q-filter to restrict the estimated bandwidth \cite{X8, X9}. This design introduces a strict trade-off in which high-frequency disturbances remain unestimable, whereas increasing the cutoff frequency inevitably amplifies the measurement noise. To better capture the coupled manipulator dynamics, a nonlinear disturbance observer (NDOB) was developed for two-link arms \cite{X10}, generalized to serial n-link manipulators with Lyapunov-based stability analysis \cite{nikoobin2009}, and then provided a systematic design free of restrictions on joint types and configurations \cite{X11}. However, the NDOB accuracy remains heavily dependent on precise model parameters. To circumvent acceleration measurement noise, momentum-based observers that leverage generalized momentum have been introduced~\cite{X12}. A comprehensive review of these architectures was provided in~\cite{X2}.

Despite these structural advancements, conventional DOBs exhibit fundamental theoretical limitations. They assume that the disturbances follow a predefined dynamic class, such as a constant or low-order polynomial, and compute the estimate as the residual between nominal expectations and measured responses. However, this premise has two critical limitations. First, unmodeled elements falling outside the assumed class, including state-dependent nonlinearities, such as Stribeck friction or high-frequency vibrations, occupy a structural blind spot. Second, the parametric uncertainties within the nominal model are incorrectly absorbed into the disturbance estimate. This estimation bias critically degrades performance during operational shifts, such as payload variations. Consequently, when multifrequency disturbances and model uncertainties act simultaneously, these fundamental boundaries persist across all analytical DOB variants.

\subsection{Learning-based Approaches and Their Integration}
\label{subsec:learning_based_approaches}

Learning-based approaches have emerged for manipulator control and disturbance estimation to overcome the limitations of model-based methods. We discuss the two threads in turn before analyzing the hybrid approaches that combine both paradigms.

\subsubsection{Learning-based Manipulator Control}
Reinforcement learning has significantly advanced manipulator control by optimizing policies without explicit dynamic models, progressing from fundamental reaching tasks~\cite{X15} to contact-rich assemblies ~\cite{X16, X17}. Despite these successes, industrial deployment remains constrained by the sensitivity of the learned policies to training data distributions. Subtle dynamic discrepancies between simulations and the physical world generally introduce variations in tracking consistency. This vulnerability is compounded by the absence of a formal mathematical framework for stability verification. Without explicit theoretical boundaries, ensuring uniform operational reliability in out-of-distribution environments is highly complex. Although sim-to-real methodologies, including domain randomization, mitigate environmental mismatches \cite{tobin2017}, they cannot formally resolve the underlying nondeterminism and the absence of stability guarantees that characterize purely learning-based architectures.

\subsubsection{Learning-based Disturbance Observation}
In addition, learning-based disturbance estimation has emerged as an approach for modeling the external dynamics directly from data without assuming predefined analytical classes. Representative methodologies include recurrent networks that forecast time-varying disturbances from historical trajectories and neural architectures designed to map complex nonlinear relationships between system states and external perturbations~\cite{X19, X21}. Although these approaches successfully capture the state-dependent nonlinearities that challenge conventional observers, they generally rely on estimating the entire disturbance profile using a single monolithic neural network. However, because this estimation capability is fundamentally bounded by the training data distribution, encountering unfamiliar disturbance profiles during operation generally amplifies estimation errors and degrades the closed-loop control performance. Furthermore, the inherent opacity of neural networks obscures the physical interpretation of learned estimates. In contrast to the explicit formulations of analytical DOBs, this lack of transparency complicates the verification process and makes it difficult to ascertain the reliability of estimation outputs.

\subsubsection{Hybrid Model-based and Learning-based Approaches}
Although analytical- and learning-based paradigms have distinct limitations, their strengths in terms of theoretical stability and empirical adaptability are inherently complementary. Consequently, hybrid architectures designed to combine these methodologies have been investigated extensively. Residual learning is a prominent framework, in which a neural policy provides additive compensation to a deterministic baseline controller \cite{silver2018, johannink2019residual, X4, X22}. In this structure, the RL component specifically targets the complex dynamics that the nominal model cannot resolve. In addition to standard residual configurations, various alternative integrations have been proposed. For instance, recent studies have augmented cost functions or predictive models of model predictive control (MPC) with learned representations~\cite{hewing2020learning, X23}, or directly coupled MPC outputs with actor-critic policies~\cite{X24}. Simultaneously, safety filters have been proposed to project learned control inputs onto certified safe sets at runtime \cite{wabersich2021}. The proposed ISS-based bound follows this principle but reduces it to a closed-form scalar clip derived from the baseline's own stability certificate, requiring no online optimization.

Although these hybrid studies have successfully validated the integration of analytical and learning-based paradigms, critical gaps remain in operational demarcation and dynamic adaptation. In most existing frameworks, the operational scopes of the deterministic baseline and neural policy significantly overlap. Because both modules simultaneously generate control inputs focusing on the identical objective of tracking error minimization, there is no formal boundary defining where the learned policy supplements the baseline or where it interferes. This structural conflation risks violating the theoretical stability of the nominal controller, forcing many contemporary architectures to rely on soft constraints or informal safety arguments rather than on rigorous mathematical bounds. Furthermore, the mechanisms that enable policies to identify and adapt to varying disturbance profiles during operation remain insufficiently investigated, particularly during abrupt environmental transitions.

To resolve these vulnerabilities, this study enforced a strict separation of operational scopes. The deterministic controller and observer handle the disturbances within their verified reliable bandwidths, assigning the unmodeled residual as the exclusive target for the RL policy. To preserve the stability margin of the baseline controller under any policy output, we derived and enforced a state-dependent action bound based on an ISS formulation.

\section{Manipulator Controller}
\label{sec:controller}

This section describes the nominal NMPC controller and model-based DOB adopted as the baseline of the proposed framework and defines the \emph{residual disturbance} that serves as the learning target for the residual RL policy. The model-based portion, shown in Fig.~\ref{fig:main_framework}, consisting of the NMPC, DOB, and plant, illustrates the cascade structure employed in this study.

\subsection{System Dynamics}
\label{subsec:dynamics}

The dynamics of an $n$-link manipulator are described in the standard Lagrangian form
\begin{equation}
M(q)\ddot{q} + C(q,\dot{q})\dot{q} + g(q) = \tau + d,
\label{eq:dynamics}
\end{equation}
where $q \in \mathbb{R}^{n}$ denotes the joint positions, $M(q)$ is the inertia matrix, $C(q,\dot{q})\dot{q}$ denotes the Coriolis and centrifugal terms, $g(q)$ is the gravitational term, $\tau \in \mathbb{R}^{n}$ is the actuator torque, and $d \in \mathbb{R}^{n}$ is the unmodeled disturbance. This study focused on a 6-DOF manipulator with $n=6$.

By defining the state vector as $x = [q^{\top}, \dot{q}^{\top}]^{\top} \in \mathbb{R}^{2n}$, the system can be written as a control-affine continuous-time dynamics.
\begin{equation}
\dot{x} = f(x, \tau) =
\begin{bmatrix}
\dot{q} \\
M^{-1}(q)\bigl(\tau - C(q,\dot{q})\dot{q} - g(q)\bigr)
\end{bmatrix}.
\label{eq:state_dynamics}
\end{equation}

\subsection{Nominal NMPC Formulation}
\label{subsec:nmpc}

In every control cycle, the nominal controller solves the following SQP-based optimization problem over a prediction horizon $N$:
\begin{equation}
\begin{aligned}
\min_{\{x_k, \tau_k, s_k\}} \quad & J = \sum_{k=0}^{N-1} \ell(x_k, \tau_k, s_k) + \ell_f(x_N) \\
\mathrm{s.t.} \quad
& x_{k+1} = x_k + \Delta t \cdot f(x_k, \tau_k), \\
& -\bar{\tau} - s_k \leq \tau_k \leq \bar{\tau} + s_k, \quad s_k \geq 0, \\
& x_0 = x(t),
\end{aligned}
\label{eq:nmpc}
\end{equation}
where $\Delta t$ is the discretization step, $\bar{\tau}$ is the torque limit, and $s_k \in \mathbb{R}^{n}$ is a soft slack variable for the torque constraint. The stage cost $\ell(\cdot)$ is expressed as follows:
\begin{equation}
\begin{aligned}
\ell(x_k, \tau_k, s_k) ={}& w_p \|p(q_k) - p^{*}\|^{2} + w_v \|\dot{q}_k\|^{2} \\
&+ w_u \|\tau_k\|^{2} + w_s \|s_k\|^{2},
\end{aligned}
\label{eq:stage_cost}
\end{equation}
where $p(q_k) \in \mathbb{R}^{3}$ is the end-effector position computed using forward kinematics, $p^{*} \in \mathbb{R}^{3}$ is the reference trajectory, and $w_p, w_v, w_u, and w_s$ are the weighting coefficients. To ensure precise trajectory tracking, the tracking term and slack penalty were heavily weighted with $w_p = 5{,}000$ and $w_s = 1{,}000$, respectively. To ensure smoothness between consecutive control inputs, an additional term $0.01 \|\tau_{k+1} - \tau_k\|^{2}$ was included. The terminal cost $\ell_f$ scaled the stage cost by a factor of five to emphasize tracking accuracy at the end of the horizon.

At each step, the first element of the resulting optimal control sequence, $\tau_{\mathrm{MPC}} = \tau_{0}^{*}$, was considered as the nominal MPC output.

\subsection{Model-based Disturbance Observer}
\label{subsec:dob}

To compensate for the disturbance $d$ that was not captured by the nominal model of the NMPC, an inverse dynamics-based DOB was integrated into a cascade structure. Using the previously applied control input $\tau_{\mathrm{cmd}}^{-}$ and the measured state $(q, \dot{q})$, the nominal torque was computed using the recursive Newton-Euler algorithm (RNEA).
\begin{equation}
\tau_{\mathrm{ID}} = M(q)\hat{\ddot{q}} + C(q,\dot{q})\dot{q} + g(q),
\label{eq:rnea}
\end{equation}
where $\hat{\ddot{q}} = (\dot{q}_t - \dot{q}_{t-1})/\Delta t$ is the joint acceleration estimated through finite differencing. The raw disturbance estimate was then obtained as the discrepancy between the nominal torque and the applied control torque.
\begin{equation}
\hat{d}_{\mathrm{raw},t} = \tau_{\mathrm{ID},t} - \tau_{\mathrm{cmd},t-1}.
\label{eq:d_raw}
\end{equation}
As $\hat{d}_{\mathrm{raw}}$ contains high-frequency noise from the measurements and numerical differentiation, a first-order IIR low-pass filter (LPF) was applied.
\begin{equation}
\hat{d}_{\mathrm{filt},t} = (1-\alpha)\hat{d}_{\mathrm{filt},t-1} + \alpha \hat{d}_{\mathrm{raw},t},
\label{eq:lpf}
\end{equation}
with a filter coefficient $\alpha = 0.2$. For the control period $\Delta t = 0.02~\mathrm{s}$ (50Hz), the cutoff frequency of the discrete LPF was determined using the following equation:
\begin{equation}
f_c = \frac{1}{2\pi \Delta t}\,\arccos\!\left(1 - \frac{\alpha^{2}}{2(1-\alpha)}\right) \approx 1.78~\mathrm{Hz}.
\label{eq:cutoff}
\end{equation}
This cutoff served as a direct reference point for designing the disturbance frequency curriculum proposed in this study (Section~\ref{sec:training}).

The nominal control law combining the NMPC and DOB is expressed as follows:
\begin{equation}
\tau_{\mathrm{cmd}} = \tau_{\mathrm{MPC}} - \hat{d}_{\mathrm{filt}}.
\label{eq:nominal_law}
\end{equation}

\subsection{Residual Disturbance}
\label{subsec:residual}

The disturbance components compensated by the nominal controller in (\ref{eq:nominal_law}) were confined within the LPF passband; the high-frequency components, nonlinear friction, and model uncertainty remained in the system without being canceled. We defined these uncompensated components as \emph{residual disturbances }.
\begin{equation}
d_{\mathrm{res}} \triangleq d_{\mathrm{true}} - \hat{d}_{\mathrm{filt}},
\label{eq:residual}
\end{equation}
where $d_{\mathrm{true}}$ is the actual disturbance in (\ref{eq:dynamics}). $d_{\mathrm{res}}$ corresponds to the signal components that the nominal DOB cannot estimate through construction and serves as the explicit learning target of the proposed residual RL framework. Section~\ref{sec:residual_rl} presents the cascade learning structure that further compensates $d_{\mathrm{res}}$ via the policy output $d_{\mathrm{rl}}$.

\begin{figure} 
    \centering
    \includegraphics[width=\linewidth]{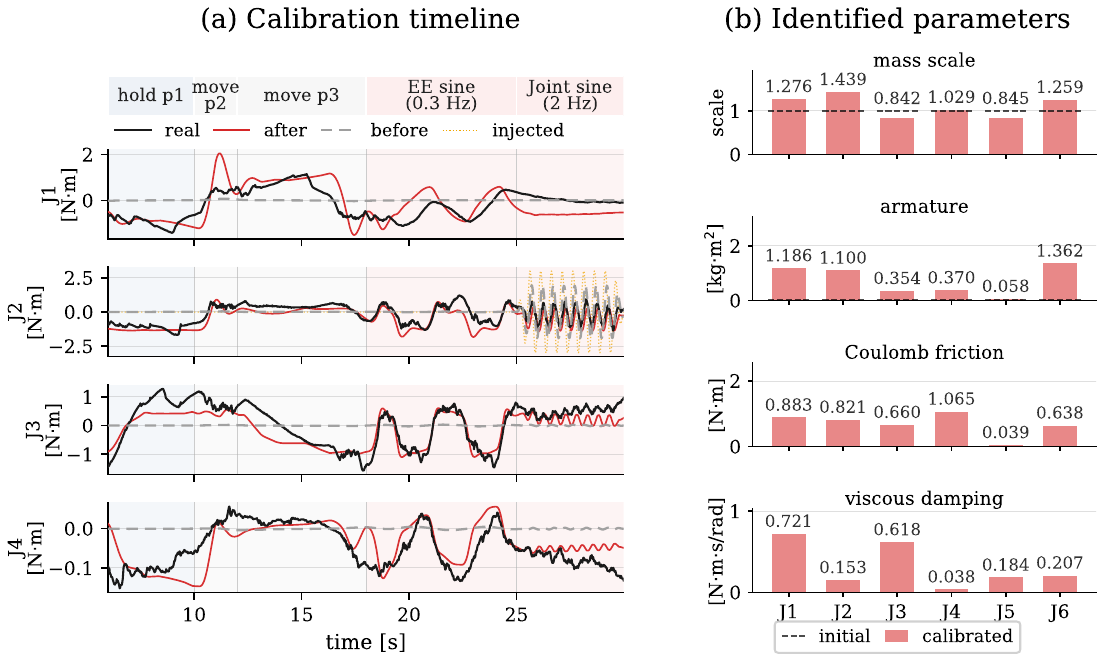} 
    \caption{System identification. (a) DOB output of the real robot compared with simulation before and after identification along the calibration trajectory. (b) Identified parameters per joint with initial values and search bounds.}
    \label{fig:sysid}
\end{figure}

\section{Residual Reinforcement Learning Framework}
\label{sec:residual_rl}

\begin{figure*}[hbt!]
    \centering
    \vspace{1.5mm}
    \includegraphics[width=\textwidth]{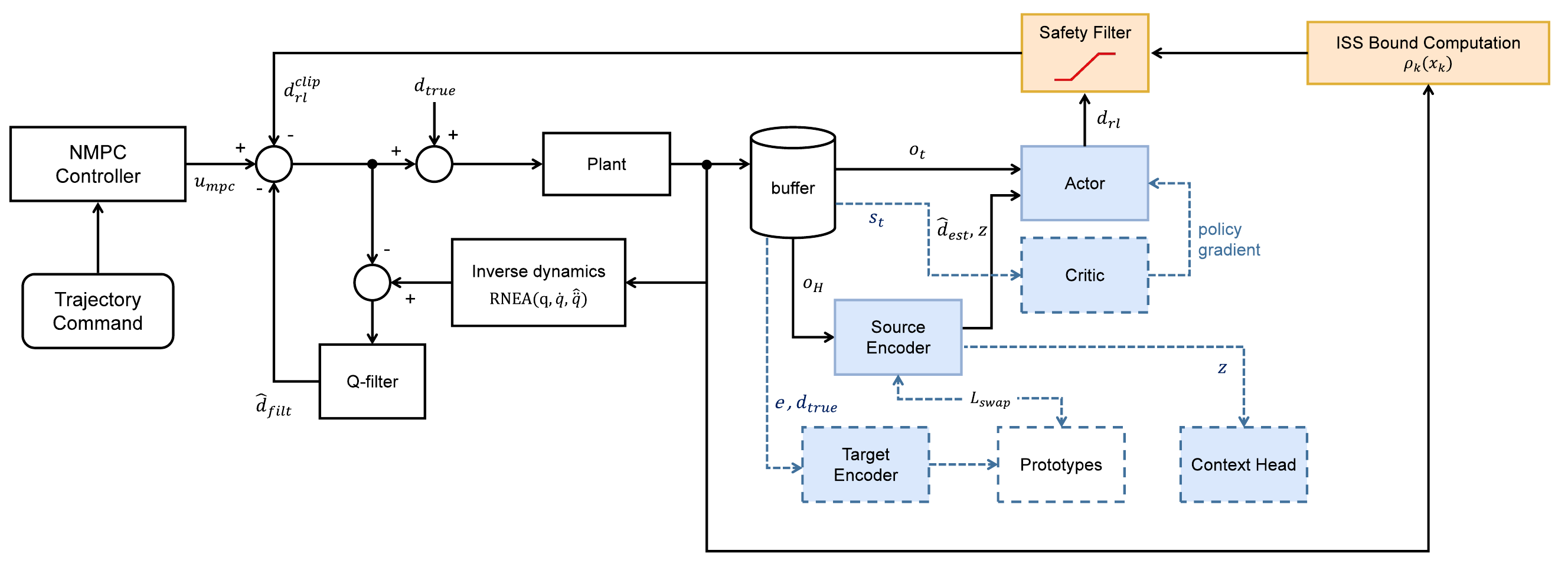}
    \caption{Training-time structure of the proposed framework. Solid blocks represent components used at deployment. Dashed blue lines (critic, target encoder, prototypes, and context head) indicate components that exist only during training.}
    \label{fig:main_framework}
\end{figure*}

\subsection{Training Environment}
\label{sec:sysid}
Fig.~\ref{fig:main_framework} illustrates the overall structure of the proposed manipulator-control framework. The RL block, which compensates for the residual disturbance, was coupled with the NMPC and DOB in a cascade configuration. Using the signals available from the existing controller as their state, the policy infers the context of the acting disturbance and learns to output the residual disturbance. The trained policy was first learned using the IsaacLab simulator and subsequently transferred to a real robot.

A prerequisite must be satisfied for this sim-to-real transfer to be successful. If the distribution of the states used as RL inputs differs substantially between the simulation and real world, the discrepancy translates directly into a sim-to-real gap; therefore, the RL states must follow similar distributions across both domains. In the simulation, the robot was instantiated from the URDF model of the Piper manipulator, yielding an idealized condition in which the plant coincides with the dynamic model used by the controller and DOB. However, a real robot does not perfectly match the URDF in terms of the joint friction, center of mass, and other properties. When the same countermass was attached and an identical trajectory command was applied to both the simulated and real robots, a comparison of the DOB outputs revealed differing trends, as shown in Fig.~\ref{fig:sysid}(a). Because the DOB output was an RL input state, the difference in trend directly induced a mismatch in the state distribution. Therefore, a stable sim-to-real transfer required the simulated robot model to approximate the real model as closely as possible.

Thus, we performed system identification using a dedicated calibration trajectory that combines several scenarios, namely holding the end effector at a fixed pose, moving it between two target points, and applying sinusoidal excitations to the end effector and joint spaces. Each scenario excites a different subset of physical parameters; therefore, the combined trajectory rendered the identification of all parameters well-posed. The trajectory was executed on a real robot while the DOB output was recorded, and an identical procedure was repeated in the simulation. A similar identification strategy was reported for legged robots \cite{bjelonic2025towards}, where the encoder bias and time delay were identified. In addition, we introduced a link mass scale that influenced the modeling of the entire robot, together with the Coulomb friction, joint armature, and viscous damping of each joint. The cross-entropy method \cite{rubinstein2004} was then used to select these parameters such that the DOB output of the simulated robot matched the recorded output, and the identified values were reflected in the simulated robot model. Fig.~\ref{fig:sysid}(a) shows a comparison of the DOB outputs before and after the identification along the calibration trajectory. Before identification, the simulated DOB output remained close to zero because the plant coincided with the nominal model described above, whereas after identification, it closely followed the trend of the real robot, confirming that the identified parameters were properly reflected in the simulation. Fig.~\ref{fig:sysid}(b) shows the identified parameters for each joint. Using this system identification, the RL policy observed the internal characteristics of the robot that were consistent with the physical platform.

\subsection{Policy Architecture}
\label{sec:policy}

The residual compensation policy was trained using PPO under an asymmetric actor-critic structure. The actor network defined policy $\pi_\phi(a_t \,|\, o_t, z_t)$ parameterized by $\phi$ and output the residual compensation action $a_t \in \mathbb{R}^6$. The observation $o_t$ consisted only of signals obtainable from the real robot.
\begin{equation}
o_t = [q_t,\, \dot{q}_t,\, p_t,\, \epsilon_t,\, a_{t-1},\,
\hat{d}_{\mathrm{filt}},\, u_{t-1}],
\label{eq:obs}
\end{equation}
where $q_t \in \mathbb{R}^6$ and $\dot{q}_t \in \mathbb{R}^6$ are the joint angles and velocities, respectively, $p_t \in \mathbb{R}^3$ is the target position, $\epsilon_t \in \mathbb{R}^3$ is the tracking error, $a_{t-1} \in \mathbb{R}^6$ is the residual compensation torque applied in the previous step, $\hat{d}_{\mathrm{filt}} \in \mathbb{R}^6$ is the DOB output, and $u_{t-1} \in \mathbb{R}^6$ is the previously applied control torque. The estimator network (Section~\ref{sec:estimator}) produced a disturbance estimate $\hat{d}_{\mathrm{est}} \in \mathbb{R}^6$ and latent context $z \in \mathbb{R}^{16}$ from the observation history, which were concatenated with $o_t$ to form the actor input. Additionally, the critic observed privileged information available only during training.
\begin{equation}
s_t = [o_t,\, d_{\mathrm{true}},\, d_{\mathrm{res}}],
\label{eq:critic_obs}
\end{equation}
where $d_{\mathrm{true}}$ is the ground-truth disturbance applied to the simulator and $d_{\mathrm{res}}$ is the residual disturbance. This asymmetric construction reduces the variance of the value estimation and aids convergence on small-magnitude learning targets such as the residual signal.

The raw actor output $a_t$ was passed through a first-order low-pass filter of the exponential moving average form and then scaled by 0.4 to yield the residual compensation torque $d_{\mathrm{rl}}$. The filter blocked the high-frequency components of the stochastic policy output from entering the torque command directly, thereby mitigating chattering on the physical actuators and actuation mismatch between training and deployment. Finally, a state-dependent bound derived from the ISS analysis was applied to $d_{\mathrm{rl}}$. The derivation of this bound and the estimation of its constants are presented in Section~\ref{sec:stability}. The manner in which latent context $z$ represents the characteristics of a disturbance is detailed in Section~\ref{sec:estimator}.

\subsection{Training Details}
\label{sec:training}

\begin{table}[t]
\centering
\caption{Disturbance sources and randomization ranges used during training. The ranges of the sinusoid frequency and payload were expanded stage by stage through the curriculum.}
\label{tab:disturbance}
\renewcommand{\arraystretch}{1.25}
\setlength{\tabcolsep}{4pt}
\footnotesize
\begin{tabular}{@{}p{0.26\linewidth} p{0.33\linewidth} p{0.33\linewidth}@{}}
\toprule
\textbf{Source} & \textbf{Mechanism} & \textbf{Range} \\
\midrule
Sinusoidal torque
& $A\sin(2\pi f t+\phi)$ on J1--J3
& $A \!\sim\! \mathcal{U}(0.2,\,2.0)$~N$\cdot$m \newline
  $f \!\sim\! \mathcal{U}(0.2,\,2.5)$~Hz (curriculum) \newline
  $\phi \!\sim\! \mathcal{U}(0,\,2\pi)$ \\
Impulse torque
& half-sine impact, random joint
& $\mathcal{U}(2,\,8)$~N$\cdot$m, $\approx\!60$~ms \\
Joint friction
& nominal profile $\times$ scale
&  $s_\mu \sim \mathcal{N}(1,\,0.2^2)$ \\
Payload
& added mass at gripper
& $m \!\sim\! \mathcal{U}(-0.1,\,3.0)$~kg \\
Sensor noise
& Gaussian on $q,\dot q$
& $\sigma_q\!=\!0.001$~rad \newline
  $\sigma_{\dot q}\!=\!0.02$~rad/s \\
\bottomrule
\end{tabular}
\end{table}

\textbf{Disturbance curriculum.}
The disturbances applied during the training consisted of four factors, as presented in Table~\ref{tab:disturbance}: joint sinusoids, end-effector payload mass, joint friction, and impulses. The joint sinusoidal disturbance was sampled per episode with an amplitude of $A \in [0.2, 2.0]$\,N$\cdot$m and frequency of $f \in [0.2, 2.5]$\,Hz and was applied simultaneously to the three proximal joints (J1–J3) with independent phases. This frequency range encompassed both the interior of the DOB low-pass passband (cutoff frequency of approximately 1.78\,Hz) and the region beyond it. Therefore, as its residual learning target, the policy experienced both the estimation error arising inside the passband and uncompensated components remaining outside it.

The joint friction was randomized by taking the joint-wise Coulomb friction and viscous damping identified through system identification (Section~\ref{sec:sysid}) as the nominal profile and multiplying it by a scale factor $s_\mu \sim \mathcal{N}(1,\,0.2^2)$ in each episode, clipped to remain positive. This design preserves the identified asymmetric friction structure across the joints and concentrates the training distribution around the physically identified values, whereas the Gaussian spread provides robustness against identification errors and friction changes during operation caused by temperature, wear, and lubrication state. Because friction identification for proximal joints has limited reliability even under dedicated excitation trajectories, the scale randomization allows the training distribution to directly absorb this identification uncertainty.

The end-effector payload mass was imposed per episode in the range of $[-0.1, 3.0]$\,kg, where the negative lower bound also included an over-modeled mass owing to the identification error in the training distribution. Impulse disturbance applied a short-duration impact torque to a random joint at regular intervals, inducing transient broadband components and training the policy to respond to momentary contact-like perturbations that persistent disturbances alone do not consider. Sinusoidal, friction, and payload parameters were registered as privileged context labels of the estimator network, allowing the network to recognize the type and magnitude of the disturbance acting on the robot (Secyion~\ref{sec:estimator}). The impulse was excluded from the labels because it is a transient event rather than a sustained regime, and its compensation was handled reactively by the policy.

The progression of the curriculum was based on the residual estimation performance. In each episode, the standalone residual of the model-based DOB, $\lVert d_{\mathrm{true}} - \hat{d}_{\mathrm{filt}} \rVert$, and the residual after RL compensation, $\lVert d_{\mathrm{true}} - (\hat{d}_{\mathrm{filt}} + d_{\mathrm{rl}}) \rVert$, were recorded, and the episode-averaged ratio of the two served as the progression indicator. When this ratio remained below the threshold of 0.75, the policy sufficiently absorbed the components that the DOB missed at the current difficulty level, and the sampling ranges of the disturbances were expanded. Conversely, when the ratio exceeded 0.92, the curriculum reverted to its previous stage. The range sets of the sinusoidal frequency and payload mass increased stage-by-stage until they reached the full randomization ranges described above, each of which contained the previous range, to ensure that low-difficulty conditions remain represented in the training distribution. Because difficulty was regulated by the residual absorption capability rather than by tracking performance, the pace of the curriculum remained aligned with the actual learning progress of the residual policy.

\textbf{Command.}
The reference command was the workspace trajectory tracking, and three trajectory families were used during training. The circle and figure-eight trajectories were sampled using a radius of $[0.05, 0.2]$\,m and a speed of $[0.5, 3.0]$\,rad/s, generating tracking conditions of varying size and speed. The random Fourier trajectory was an aperiodic trajectory synthesized from three sinusoidal components with random coefficients, which prevented overfitting to fixed-shape trajectories and enlarged the visited region of the state space such that the residual compensation performance was maintained under changes in the trajectory shape. In addition, a time-warp perturbation was applied to the traversal speed in most episodes, inducing acceleration and deceleration along the trajectory to avoid overfitting the policy to constant-velocity tracking and maintaining its residual compensation performance through speed transitions.

\begin{table}[t]
\centering
\caption{Reward function components and weights.}
\label{tab:reward}
\renewcommand{\arraystretch}{1.35}
\setlength{\tabcolsep}{5pt}
\footnotesize
\begin{tabular}{@{}l l c@{}}
\toprule
\textbf{Term} & \textbf{Expression} & \textbf{Weight} \\
\midrule
Residual matching
& $0.1 \,/\, \big(\lVert d_{\mathrm{rl},t} - d_{\mathrm{res},t} \rVert_1 + 0.1\big)$
& $3.5$ \\
Compensation smoothness
& $\lVert \Delta(\hat{d}_{\mathrm{filt},t} + d_{\mathrm{rl},t}) \rVert^2$
& $-0.3$ \\
EE smoothness
& $\exp\!\big(-\lVert \ddot{p}_{ee,t} \rVert^2 /0.5\big)$
& $0.3$ \\
Action magnitude
& $\lVert a_t \rVert^2$
& $-0.01$ \\
Joint acceleration
& $\lVert \ddot{q}_t \rVert^2$
& $-2.5\!\times\!10^{-5}$ \\
\bottomrule
\end{tabular}
\end{table}

\textbf{Reward functions.}
The reward terms are summarized in Table~\ref{tab:reward}. The core term was a residual matching term that directly rewarded the degree to which the policy output cancelled the actual residual disturbance, restricting the learning target to the components that the DOB failed to compensate for, thereby preventing role overlap with the baseline. In addition, penalties on the action magnitude, joint acceleration, and rate of the combined compensation torque suppressed large and rapidly changing outputs. A bounded end-effector smoothness reward further enhanced smooth motion, which was induced together with the low-pass filter described in Section~\ref{sec:policy} and a smooth compensation torque.


\begin{table}[t]
\centering
\caption{Composition of the privileged disturbance context $e \in \mathbb{R}^{8}$.}
\label{tab:context}
\renewcommand{\arraystretch}{1.2}
\footnotesize
\begin{tabular}{@{}lll@{}}
\toprule
\textbf{Channel} & \textbf{Meaning} & \textbf{Source} \\
\midrule
$A_{1..3}$ & sinusoid amplitudes (J1--J3) & episode parameters \\
$f_{1..3}$ & sinusoid frequencies (J1--J3) & episode parameters \\
$m$ & end-effector payload offset & mass randomization \\
$s_{\mu}$ & joint friction scale & friction randomization \\
\bottomrule
\end{tabular}
\end{table}

\subsection{Estimator Network}
\label{sec:estimator}

\begin{figure} 
    \centering
    \includegraphics[width=\linewidth]{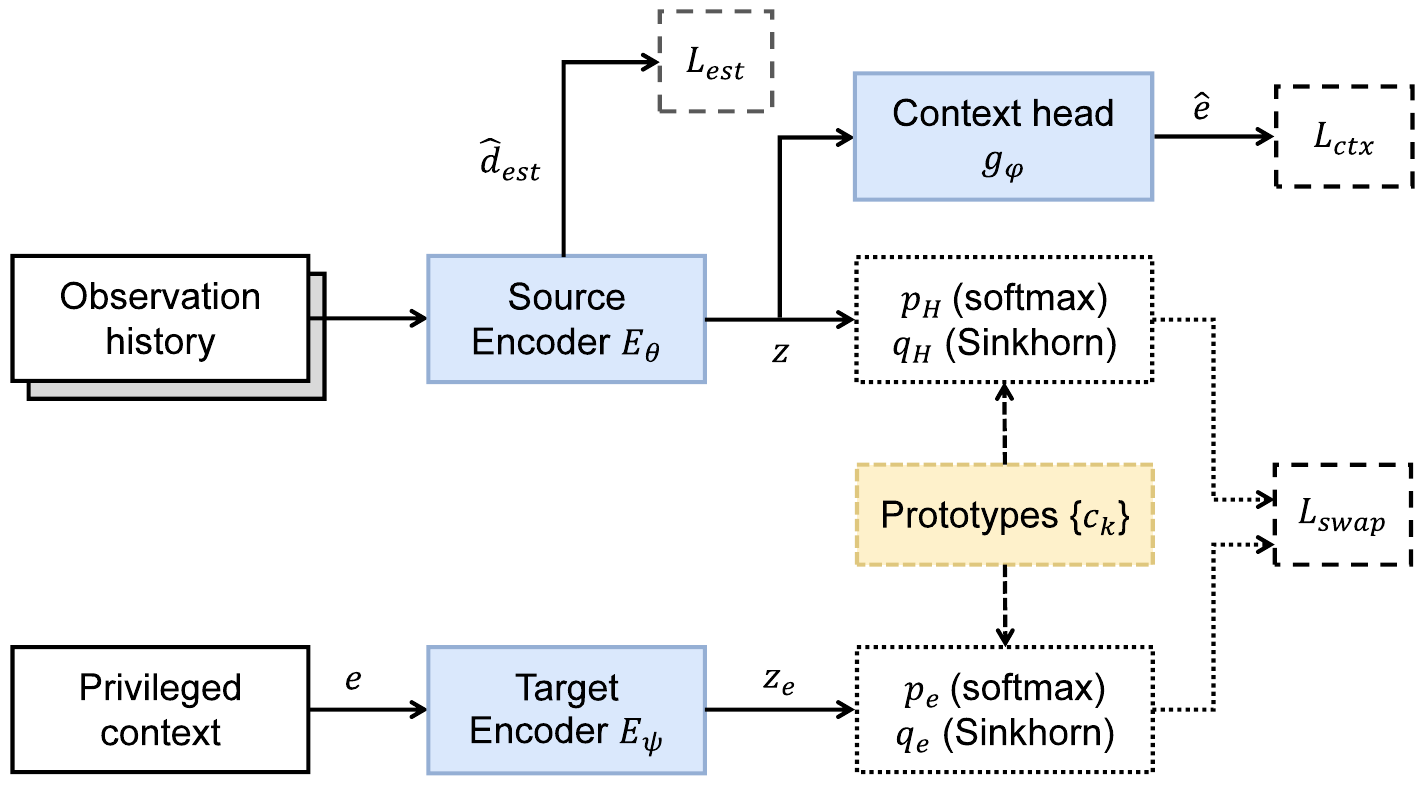} 
    \caption{Estimator network. The source encoder maps the observation history to a disturbance estimate and latent context, and is aligned with a target encoder that embeds the privileged context through prototype assignment and context regression.}
    \label{fig:estimator}
\end{figure}

Residual disturbance is a mixture of heterogeneous sources, and each source requires a different compensation strategy. Periodic excitations require predictive, phase-leading compensation; static loads require a constant offset; and impulsive events require a damping response that suppresses the ensuing transient. If the policy receives only instantaneous observations, it must reinfer the acting disturbance from scratch at every step, which hampers learning and slows adaptation whenever the disturbance changes. To resolve this, the estimator network learned a latent representation that summarized the type and intensity of the acting disturbance based on the observation history. This summary is referred to as the \emph{disturbance regime}. The representation should be similar for histories generated under the same regime, distinct across regimes, and invariant to regime-irrelevant factors, such as the trajectory phase or the instantaneous posture of the robot.

As shown in Fig.~\ref{fig:estimator}, the estimator comprised two encoders during training. The source encoder $E_\theta$ mapped a 1.0\,s (50-step) observation history $o_H$ to a disturbance estimate and latent context, and was the only component used during deployment. The target encoder $E_\psi$ was used exclusively during training. It received the privileged disturbance context $e$ (Table~\ref{tab:context}), known only to the simulator, and embedded it in the same latent space.
\begin{equation}
(\hat{d}_{\mathrm{est}},\, z) = E_\theta(o_H), \qquad
z_e = E_\psi(e).
\label{eq:encoders}
\end{equation}

The estimate $\hat{d}_{\mathrm{est}}$ was trained through direct regression against the true disturbance $\mathcal{L}_{\mathrm{est}} = \lVert \hat{d}_{\mathrm{est}} - d_{\mathrm{true}} \rVert^{2}$, providing the policy with an immediately usable feed-forward signal. However, this loss provides no incentive for the latent representation to organize itself by regime because the two histories produced by different disturbances may be mapped to nearby codes as long as their instantaneous residuals agree. Therefore, we introduced two auxiliary losses that impose a regime structure on the latent space, the sum of which is referred to as the regime loss, $\mathcal{L}_{\mathrm{regime}} = \mathcal{L}_{\mathrm{swap}} + \mathcal{L}_{\mathrm{ctx}}$.

The first loss was aligned with the \emph{discrete} cluster structure. We defined $K = 32$ learnable prototypes $\{c_k\}$ in the latent space and mapped each representation to a soft assignment over the prototypes. For representation $z$, the predicted distribution $\mathbf{p}$ and alignment target $\mathbf{q}$ are defined as follows:
\begin{equation}
p^{(k)} = \frac{\exp(z^{\top} c_k / T)}
               {\sum_{k'} \exp(z^{\top} c_{k'} / T)},
\qquad
\mathbf{q} = \mathrm{Sinkhorn}\!\left(\mathbf{Z}^{\top}\mathbf{C}\right),
\label{eq:proto}
\end{equation}
where $T$ is the temperature and $\mathrm{Sinkhorn}(\cdot)$ is the optimal transport normalization that enforces balanced cluster usage over the batch assignment matrix \cite{cuturi}. This normalization prevents representation collapse, in which all inputs concentrate on a few clusters, and drives the $K$ prototypes to partition the regime space evenly. The training objective was swapped prediction \cite{X5}. The encoder that has seen only history must point to the same cluster as the encoder that has seen the ground-truth context, and vice versa, and is expressed as follows:
\begin{equation}
\mathcal{L}_{\mathrm{swap}} =
-\tfrac{1}{2}\,\mathbb{E}\!\left[
\mathbf{q}_{e}^{\top}\log\mathbf{p}_{H} +
\mathbf{q}_{H}^{\top}\log\mathbf{p}_{e}\right].
\label{eq:swap}
\end{equation}
The learning signal agreed with the cluster assignment rather than with the value of $e$. The regimes were grouped according to the number of clusters that were not specified a priori but emerged during training; thus, the continuous disturbance-parameter space was automatically organized into a finite set of regime clusters.

The second loss recovered \emph{continuous} parameters. Cluster membership indicated the type of regime; however, the information that the policy required within a regime, such as the amplitude of the excitation or the weight of the payload, was continuous. A lightweight context head $g_\phi$ reconstructed context directly from a latent representation $\hat{e} = g_\phi(z)$. The eight channels of $e$ differed significantly in how they were observed throughout history. For instance, the friction scale is barely observable while the joints move slowly because the friction torque manifests only through motion. Therefore, regressing all channels with equal weights forced the encoder to chase targets that history could not explain, and the resulting gradients acted as noise that corrupted the regime structure of the latent space. Instead, each channel was weighted by learnable uncertainty $\sigma_i$ \cite{kendall2018}, yielding the following equation:
\begin{equation}
\mathcal{L}_{\mathrm{ctx}} =
\sum_{i=1}^{8} \frac{(\hat{e}_i - e_i)^2}{2\sigma_i^2} + \log \sigma_i .
\label{eq:ctx}
\end{equation}
For a channel with a large error, minimizing the loss increases $\sigma_i$, which attenuates its gradient by a factor of $1/2\sigma_i^2$, whereas the $\log\sigma_i$ term prevents the trivial solution of inflating every uncertainty. The estimator learns not only the context values but also how much each channel should be trusted, concentrating the latent capacity on the channels that history can explain without any manual per-channel weight tuning.

Table~\ref{tab:context} lists the composition of the privileged context $e$ used as the alignment target. Prior estimators in legged locomotion \cite{himloco} adopted a self-supervised construction that aligns history with the next observation $o_{t+1}$, which is effective because environmental factors such as terrain and friction leave traces in the next state. However, in an active disturbance-rejection architecture, this premise is not applicable. The more effectively NMPC and DOB cancel the disturbance, the more thoroughly their traces are erased from the next state. A latent trained with $o_{t+1}$-alignment collapses all disturbance conditions onto a single-phase orbit and carries no discriminative power. Therefore, we replaced the alignment target with a privileged regime descriptor $e$. Because $e$ is constant or varies slowly within an episode, histories that differ only in phase under the same regime were forced into the same assignment, whereas histories from different regimes were forced apart, thereby realizing, at the level of the objective function, the regime-sensitive and phase-invariant representation demanded at the outset.

The trained estimator provided a policy with information on two levels. The estimate $\hat{d}_{\mathrm{est}}$ supplied an instantaneous joint-wise disturbance signal, whereas $z$ carried macroscopic information of the current regime. As regimes are represented as a cluster structure, a disturbance transition conditions the policy simply through the movement of $z$ to the corresponding cluster, thereby enabling rapid adaptation without a separate re-identification procedure. Similarly, the disturbance-free quiet condition is represented by an explicit cluster corresponding to zero context, giving the policy a reference point at which it refrains from unnecessary intervention. The target encoder, prototypes, and context head exist only to shape the latent space. During deployment, only a single forward pass of the source encoder was added; therefore, the framework incurred no practical increase in inference cost, while yielding faster training convergence and reduced transient error during disturbance transitions.

\section{Stability Analysis}
\label{sec:stability}

The NMPC--DOB base system admits standard stability analysis, whereas the residual RL policy is a stochastic neural network for which no closed-form certificates exist. Therefore, we proceeded with two steps. We first demonstrated that the NMPC--DOB base system is ISS \cite{sontag2008} with respect to the residual disturbance $d_{\mathrm{res}}$, and then derived from this certificate a hard bound on the RL action; thus, the policy may learn freely within the bound while retaining a certified error envelope of the base system.

\subsection{ISS of the Base System}
\label{sec:iss_base}

We let the tracking-error state be as follows:
\begin{equation}
\boldsymbol{x}_k=\big[(\boldsymbol{q}_k-\boldsymbol{q}_k^{\mathrm{ref}})^\top,\;
(\dot{\boldsymbol{q}}_k-\dot{\boldsymbol{q}}_k^{\mathrm{ref}})^\top\big]^\top
\in\mathbb{R}^{12},
\label{eq:err_state}
\end{equation}
where $\boldsymbol{x}_k=0$ denotes perfect tracking. For national convenience, the bold $\boldsymbol{x}_k$ hereafter denotes this tracking-error state and is distinct from the plant state $x$ described in Section~\ref{sec:controller}. We adopted the NMPC optimal cost $V_N^\ast(\boldsymbol{x})$ as a Lyapunov candidate for the entire system and assumed three properties in a local region $\mathcal{X}$ containing the origin. \textbf{(A1)} Class-$\mathcal{K}_\infty$ functions bound the cost as $\alpha_1(\|\boldsymbol{x}\|)\le V_N^\ast(\boldsymbol{x}) \le\alpha_2(\|\boldsymbol{x}\|)$. \textbf{(A2)} The stage cost is lower-bounded as $\ell(\boldsymbol{x},\kappa_N(\boldsymbol{x}))\ge c_x\|\boldsymbol{x}\|^2$ for some $c_x>0$. \textbf{(A3)} The optimal cost is locally Lipschitz, $|V_N^\ast(\boldsymbol{x}+\delta)-V_N^\ast(\boldsymbol{x})| \le\gamma_0\|\delta\|$ for some $\gamma_0>0$. These were mild. (A1) follows from the positive-definiteness and radial unboundedness of the cost, (A2) holds by construction for the quadratic stage cost in the NMPC formulation, and (A3) is the standard local Lipschitz continuity of the optimal cost.

These properties are inherently local. Region $\mathcal{X}$ is assumed to remain away from kinematic singularities, where the inertia matrix is uniformly positive definite and the finite-horizon optimal cost is smooth; thus, the assumptions hold despite the nonlinearity of the manipulator dynamics. Constructing such a region explicitly is intractable for a 6-DOF nonlinear system, and following the standard practice in terminal-cost NMPC stability analysis, we assumed its existence rather than computing it. Therefore, the stability guarantees are conditional. Provided that the baseline NMPC--DOB system satisfies (A1)–(A3) in its operating region, the residual RL policy retains the certified error envelope of the base system. The empirical estimation of $c_x$ described later serves as a data-driven confirmation that the decreasing condition indeed holds in the actual operating region, instead of an analytic construction of $\mathcal{X}$.

First, we considered a base system with RL disabled ($d_{\mathrm{rl}}=0$). Under (A1) and (A2) and the standard NMPC terminal condition \cite{X1}, the optimal cost decreases monotonically along the disturbance-free closed loop as follows:
\begin{equation}
V_N^\ast(\boldsymbol{x}_{k+1}^{\mathrm{nom}})-V_N^\ast(\boldsymbol{x}_k)
\le-c_x\|\boldsymbol{x}_k\|^2,
\\
\boldsymbol{x}_{k+1}^{\mathrm{nom}}
= f(\boldsymbol{x}_k,\kappa_N(\boldsymbol{x}_k)).
\label{eq:nominal_descent}
\end{equation}
The constant $c_x$ quantifies the extent to which the controller pulls the error toward the origin.

Because the DOB compensation $-\hat{d}_{\mathrm{filt},k}$ is embedded in the applied input, the true disturbance $d_{\mathrm{true},k}$ enters the nominal closed loop only through its uncompensated remainder $d_{\mathrm{res},k} =d_{\mathrm{true},k}-\hat{d}_{\mathrm{filt},k}$. Thus, the actual successor state shifts to $\boldsymbol{x}_{k+1}=\boldsymbol{x}_{k+1}^{\mathrm{nom}} +d_{\mathrm{res},k}$. Adding and subtracting $V_N^\ast(\boldsymbol{x}_{k+1}^{\mathrm{nom}})$ separates the one-step cost change into disturbance-induced perturbation and nominal descent. Bounding the perturbation by the Lipschitz property (A3) and the descent by \eqref{eq:nominal_descent} yields the ISS bound of the base system.
\begin{equation}
V_N^\ast(\boldsymbol{x}_{k+1})-V_N^\ast(\boldsymbol{x}_k)
\le-c_x\|\boldsymbol{x}_k\|^2
+\gamma_0\|d_{\mathrm{res},k}\|.
\label{eq:iss_bound}
\end{equation}
The negative quadratic term is the dissipation supplied by the controller, and the linear term is the cost injected by the residual. Their balance determines the steady-state tracking radius, and this competition is precisely exploited by the action bound below.

\subsection{State-Dependent Stability Filter}
\label{sec:iss_filter}

We fixed the radius $r > 0$ within which the tracking error should settle. In the ISS bound of \eqref{eq:iss_bound}, the cost ceases to decrease when the injection term cancels the dissipation supplied by the controller. For this balance to occur no closer to the origin than $\|\boldsymbol{x}_k\| = r$, the residual must satisfy $\gamma_0\|d_{\mathrm{res},k}\| \le c_x r^2$. This defines the \emph{admissible residual budget}, where the largest residual of the base system can be absorbed while maintaining the tracking error within radius $r$.
\begin{equation}
\|d_{\mathrm{res},k}\| \le \bar{d}_{\mathrm{res}}
\triangleq \frac{c_x r^2}{\gamma_0}.
\label{eq:budget}
\end{equation}

Activating the RL policy shifts the disturbance seen by the base system to the effective residual $d_{\mathrm{eff},k} = d_{\mathrm{res},k} - d_{\mathrm{rl},k}$; thus, the injection term in \eqref{eq:iss_bound} becomes $\gamma_0\|d_{\mathrm{res},k} - d_{\mathrm{rl},k}\|$. To maintain the validity of the base ISS certificate while retaining the same constants, the policy should not enlarge the residual it is intended to cancel.
\begin{equation}
\|d_{\mathrm{res},k} - d_{\mathrm{rl},k}\| \le \|d_{\mathrm{res},k}\|.
\label{eq:no_enlarge}
\end{equation}

The purpose of a stability-preserving action bound is to ensure that the closed loop remains contracting toward the neighborhood of the origin, regardless of the policy outputs. It is sufficient for $V_N^\ast$ to decrease whenever the error satisfies $\|\boldsymbol{x}_k\| > r$, which is guaranteed by the condition
\begin{equation}
\gamma_0\|d_{\mathrm{eff},k}\| \le c_x\|\boldsymbol{x}_k\|^2 .
\label{eq:decrease_cond}
\end{equation}
Because the effective residual involves the unmeasurable true residual $d_{\mathrm{res},k}$, we bound it through the triangle inequality, which corresponds to the worst-case alignment in which the policy output points exactly opposite to the residual.
\begin{equation}
\|d_{\mathrm{eff},k}\|
\le \|d_{\mathrm{res},k}\| + \|d_{\mathrm{rl},k}\|
\le \bar{d}_{\mathrm{res}} + \|d_{\mathrm{rl},k}\|.
\label{eq:eff_bound}
\end{equation}
Substituting \eqref{eq:eff_bound} into the decrease condition \eqref{eq:decrease_cond} and using $\bar{d}_{\mathrm{res}} = c_x r^2/\gamma_0$ removes the dependence on the unmeasurable $d_{\mathrm{res},k}$ entirely, leaving a bound on the action magnitude that depends only on the current tracking error.
\begin{equation}
\|d_{\mathrm{rl},k}\| \le \rho_k
\triangleq \frac{c_x}{\gamma_0}
\left(\|\boldsymbol{x}_k\|^2 - r^2\right)_{+},
\label{eq:rho}
\end{equation}
where $(\cdot)_{+} = \max(\cdot, 0)$. When the error lies inside the target radius $\|\boldsymbol{x}_k\| \le r$, the bound vanishes, and the policy output is suppressed. Thus, the filter acts as a deadband to prevent unnecessary intervention and chattering near the target. In contrast to the constant bound obtained by fixing the worst-case state at $\|\boldsymbol{x}_k\| = r$, the bound $\rho_k$ reflects the dissipation margin $c_x\|\boldsymbol{x}_k\|^2$ available at each step. Consequently, the admissible authority increases in proportion to $\|\boldsymbol{x}_k\|^2$ in the transient regimes, where $\|\boldsymbol{x}_k\| \gg r$ and the policy can compensate for the residual disturbance and tighten only near the target.

We enforced this bound as a direction-preserving radial saturation with a state-dependent radius placed between the actor network and the joint command.
\begin{equation}
d_{\mathrm{rl},k}^{\mathrm{clip}}
= d_{\mathrm{rl},k} \cdot
\min\!\left(1,\; \frac{\rho_k}{\|d_{\mathrm{rl},k}\|}\right).
\label{eq:clip}
\end{equation}
The clipping operation depends only on fixed constants $(c_x, \gamma_0, r)$, the current error norm, and action magnitude. Therefore, it can be reduced to a scalar computation that does not require optimization, with negligible runtime cost.

This design provides a firm contraction guarantee in the worst-case scenario. When the bound $\rho_k$ is enforced, even if the policy output points exactly opposite the residual, the effective residual is limited to
\begin{equation}
\|d_{\mathrm{eff},k}\|
\le \bar{d}_{\mathrm{res}} + \rho_k
= \frac{c_x}{\gamma_0}r^2
+ \frac{c_x}{\gamma_0}\left(\|\boldsymbol{x}_k\|^2 - r^2\right)
= \frac{c_x}{\gamma_0}\|\boldsymbol{x}_k\|^2,
\label{eq:eff_worst}
\end{equation}
which satisfies
$\gamma_0\|d_{\mathrm{eff},k}\| \le c_x\|\boldsymbol{x}_k\|^2$, Hence
\begin{equation}
V_N^\ast(\boldsymbol{x}_{k+1}) - V_N^\ast(\boldsymbol{x}_k)
\le -c_x\|\boldsymbol{x}_k\|^2 + \gamma_0\|d_{\mathrm{eff},k}\|
\le 0
\label{eq:monotone}
\end{equation}
satisfies an arbitrary policy output. Therefore, $V_N^\ast$ does not increase in every state with $\|\boldsymbol{x}_k\| > r$, and when combined with (A1), the tracking error contracts monotonically to the sublevel set corresponding to radius $r$. Even if the policy fails arbitrarily, the error contracts with the certified envelope of radius $r$. In normal operation, when the policy adequately cancels the residual, it settles inside the envelope more quickly.

\begin{figure} 
    \centering
    \includegraphics[width=\linewidth]{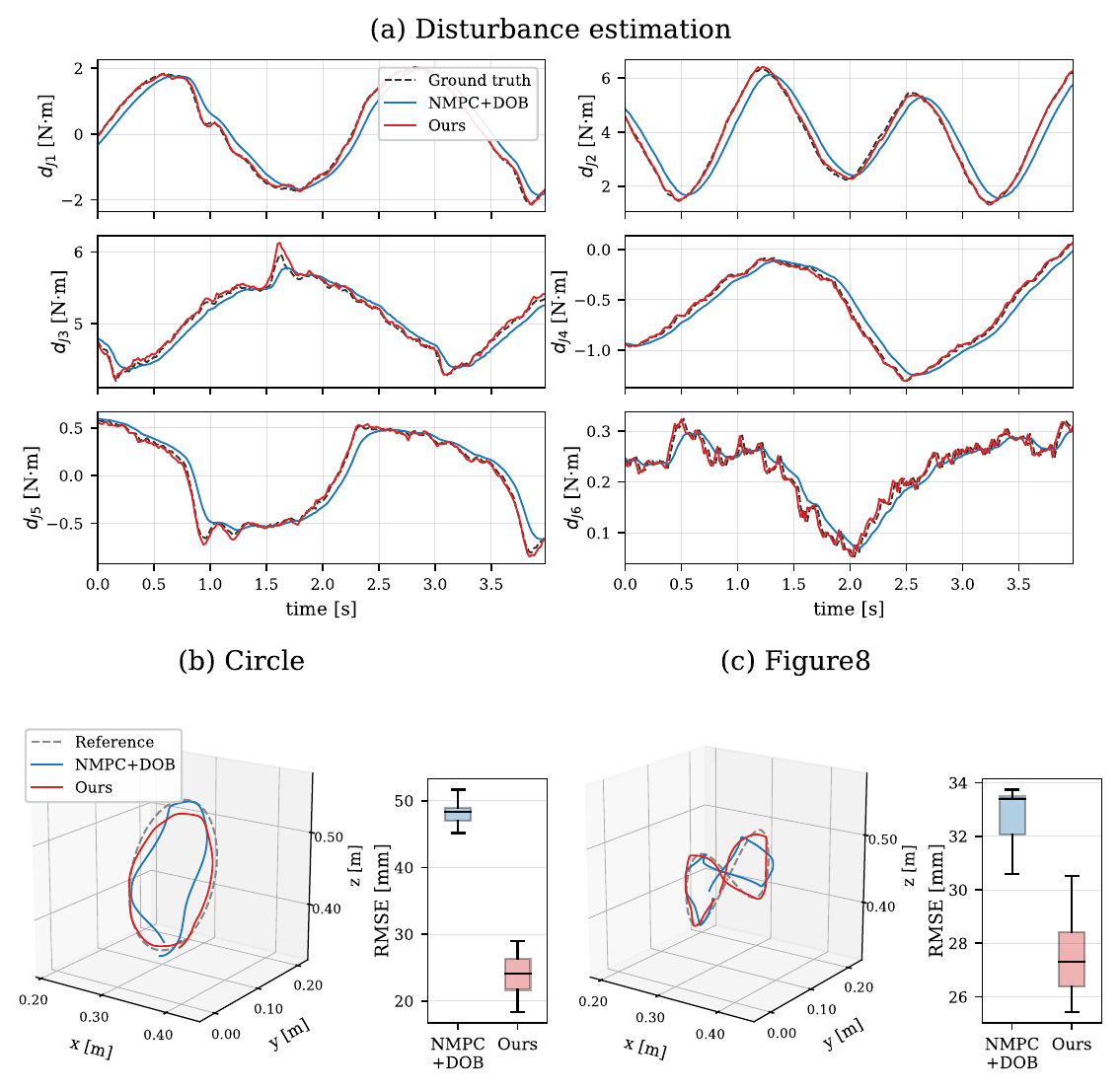} 
    \caption{Simulation results. (a) Joint-wise disturbance estimation in IsaacLab. (b) and (c) Zero-shot tracking in Gazebo under compound disturbances on circular and figure-eight trajectories, respectively, with per-cycle RMSE.}
    \label{fig:dist_est}
\end{figure}

The bound \eqref{eq:rho} is exact but conservative for two reasons. First, the certificate requires constants that remain uniform throughout the operating region. Therefore, the ratio $c_x/\gamma_0$ pairs a conservative percentile of the dissipation with a conservative percentile of the sensitivity, and the resulting uniform ratio charges every state the price of the least favorable one, even though the states that realize the weakest dissipation and those that realize the highest sensitivity are generally distinct. Second, the triangle inequality \eqref{eq:eff_bound} admits the worst-case orientation, in which the policy output points exactly opposite to the residual, whereas the policy is trained by the residual matching reward to cancel the residual; thus, its output is driven toward the opposite extreme of the alignment. 
Under these stacked worst cases, evaluating \eqref{eq:rho} at a representative transient error $\|\boldsymbol{x}_k\| \approx 0.1$ with the estimated constants yields $\rho_k$ of the order of $0.04$\,N$\cdot$m, whereas the residuals to be compensated reach the N$\cdot$m scale (Fig.~\ref{fig:dist_est}(a)), which passes only a small percentage of the required authority and certifies a closed loop in which learned compensation is never permitted to act. Relaxing the constants would invalidate the decrease inequality \eqref{eq:iss_bound} on the states represented by the percentiles. Thus, we instead maintained \eqref{eq:rho} intact as the anchor of the analysis and exposed the entire relaxation through a single factor.

Therefore, in the deployment, we used the following practical form:
\begin{equation}
\rho_k = \min\!\left(\kappa\,\frac{c_x}{\gamma_0}
\left(\|\boldsymbol{x}_k\|^2 - r^2\right)_{+},\; \rho_{\max}\right),
\label{eq:rho_practical}
\end{equation}
where $\kappa \ge 1$ restores the authority removed by stacked worst-case constructions and $\rho_{\max}$ captures the authority in the large-error regime. Price is both finite and explicit. Because $\rho_k \le \rho_{\max}$, the ISS argument described in Section~\ref{sec:iss_base} remains valid as the injection term increases by at most $\gamma_0\rho_{\max}$, and the certified envelope increases from $r$ to $r' = \sqrt{r^2 + \gamma_0\rho_{\max}/c_x}$, retaining the contraction \eqref{eq:monotone} outside $r'$ and the boundedness inside. Therefore, the choice of $\kappa$ trades off the tightness of the certified envelope against the compensation authority in transients. We set $\kappa = 3.5$ and empirically validated this choice as described in Section~\ref{sec:exp_ablation}, where the deployed bound was found to be adequate in both directions. It remained above the ground-truth residual magnitude such that legitimate compensation was not suppressed; however, it tracked it with sufficient accuracy that no authority was granted beyond the residual demands.

It should be noted that this bound ensures that the policy does not degrade the closed loop, and that the error contracts to the certified envelope. If the actual residual $\|d_{\mathrm{res},k}\|$ exceeds the reserved budget $\bar{d}_{\mathrm{res}}$, the final settling radius is determined not by $r$ but by the effective residual, namely $\sqrt{\gamma_0\|d_{\mathrm{eff},k}\|/c_x}$. Therefore, in this framework, $r$ is interpreted not as a precision target under disturbance-free operation but as the deadband radius at which policy intervention ceases, and the role of the RL policy is to reduce the effective residual $\|d_{\mathrm{eff},k}\|$ to bring this settling radius close to the target $r$.

Finally, the bound requires three constants $(c_x, \gamma_0, r)$, all of which were estimated through simulation. We estimated $c_x$ as a conservative low percentile of the per-step cost decrease ratio $\left[V_N^\ast(\boldsymbol{x}_k) - V_N^\ast(\boldsymbol{x}_{k+1})\right]/\|\boldsymbol{x}_k\|^2$ recorded under disturbance-free operation, and $\gamma_0$ as a high percentile of $\left[-\Delta V_k + c_x\|\boldsymbol{x}_k\|^2\right]/ \|d_{\mathrm{res},k}\|$ obtained by rearranging \eqref{eq:iss_bound}. The denominator of the latter requires a clean ground-truth residual, which can only be obtained through simulation based on inverse dynamics evaluated using physics-engine ground-truth acceleration. 
Because the system identification procedure aligns the simulated dynamics with those of a real robot, the simulated rollouts represent the same operating region and yield the same distribution of the state and cost samples as the real platform. The estimates are therefore representative of the real operating regime, even though the ground-truth residual is not measured on the hardware. Because $V_N^\ast$ is mathematically identical in both domains with the same NMPC weights, horizon, and sample time, they are transferred directly to the real robot. Radius $r$ is the only design variable that sets both the deadband at which policy intervention ceases and the target settling radius. As the reserved budget $\bar{d}_{\mathrm{res}} = c_x r^2/\gamma_0$ scales with $r^2$, $r$ is the dominant design variable governing both the compensation authority granted to the policy and the target-settling accuracy, and is matched to the steady-state accuracy achieved by the base system. In our implementation, the estimation yielded $c_x = 9.65$ and $\gamma_0 = 2.02$ (10th and 90th percentiles, respectively), where we set $r = 0.05$, $\kappa = 3.5$, $\rho_{\max} = 3.0$.

\section{Experiments}
\label{sec:experiments}

The proposed framework was validated on three levels. Section~\ref{sec:exp_sim} discusses the evaluation of the disturbance estimation and tracking in the simulation. Section~\ref{sec:exp_ablation} discusses the verification of the stability filter and isolation of the contribution of each component through ablation. Section~\ref{sec:exp_real} confirms the results for real hardware, including a disturbance source that was not observed during training.

\subsection{Disturbance Estimation and Tracking in Simulation}
\label{sec:exp_sim}
Fig.~\ref{fig:dist_est}(a) shows the jointwise disturbance estimation results under sinusoidal disturbances in the IsaacLab environment. Owing to the structural limitation of its low-pass filter, the model-based baseline (NMPC+DOB in the figures) exhibited a consistent phase lag and amplitude deficit relative to the ground-truth disturbance and failed to estimate the high-frequency components beyond its passband. By contrast, the proposed method (our method) combined with residual RL removed the phase lag and tracked even high-frequency fluctuations while faithfully recovering the small-signal components of the wrist joints (J5 and J6) dominated by friction and gravity residuals. Over the selected window, the joint-averaged estimation error was reduced by approximately 79.1\% compared to the DOB alone. This demonstrates that the learned residual precisely complements the structural limitation of the DOB, namely, its filter lag.

Fig.~\ref{fig:dist_est}(b) and (c) show the trajectory tracking results after transferring the policy trained in IsaacLab to the Gazebo environment without any parameter adjustments under the simultaneous action of three compound disturbances: sinusoids, joint friction, and a countermass. The difference in the disturbance estimation capability translated directly into a tracking error. The controller with the model-based DOB alone failed to absorb the residual compound disturbances and exhibited a large tracking error, whereas the proposed method reduced the per-cycle RMSE from 48.0\,mm to 24.4\,mm on the circular trajectory (49.2\% reduction) and from 32.7\,mm to 27.4\,mm on the figure-eight trajectory (16.1\% reduction). This is consistent with the design hypothesis of the framework that improved residual estimation accuracy leads to improved tracking performance. It shows that the performance gain is preserved in a simulator with a physics engine that is different from the training environment.

\begin{figure} 
    \centering
    \includegraphics[width=\linewidth]{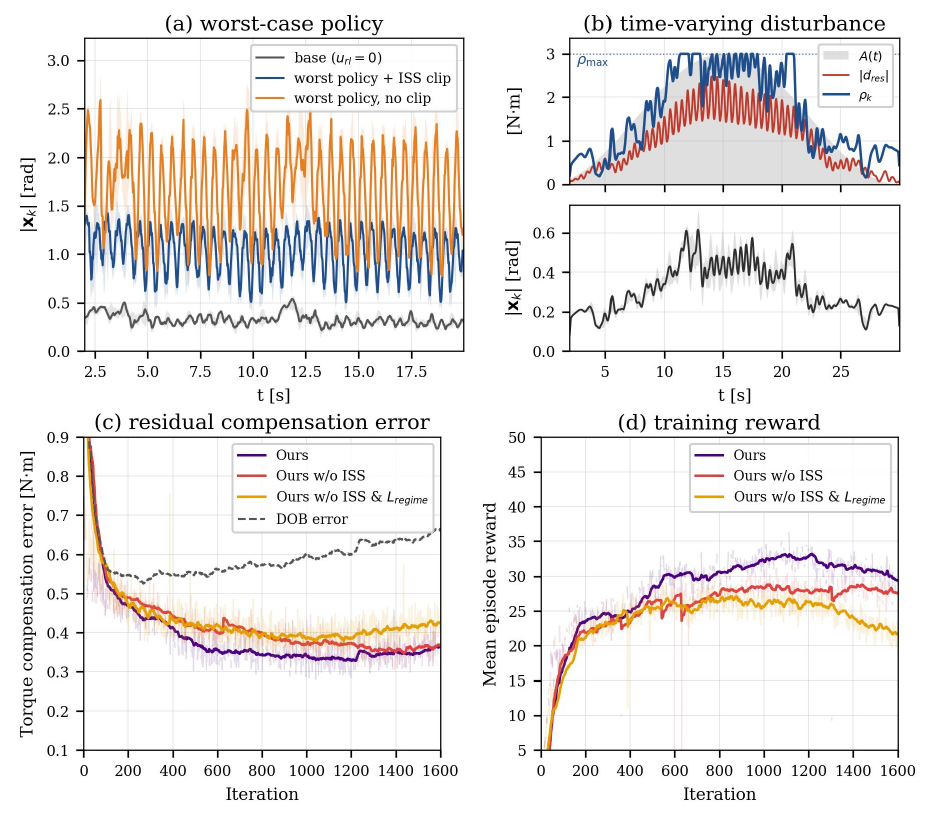} 
    \caption{Stability filter and ablation. (a) State error under an adversarial policy with and without the ISS bound. (b) Admissible bound $\rho_k$ under a time-varying disturbance. (c) and (d) Residual compensation error and episode reward during training, respectively; ablations are cumulative.}
    \label{fig:iss}
\end{figure}

\subsection{Stability Filter and Ablation Study}
\label{sec:exp_ablation} 
Fig.~\ref{fig:iss}(a) and (b) show the validation results of the stability guarantee of the state-dependent bound $\rho_k$ described in Section~\ref{sec:stability}. To construct the worst-case condition, we replaced the learned policy with an adversarial policy that outputs the maximum admissible torque in the same direction as the applied disturbance and measured the state error norm $\lVert \boldsymbol{x}_k \rVert$. Fig.~\ref{fig:iss}(a), without the bound (orange), shows that the state error was strongly amplified relative to the baseline without RL intervention (base, $u_{rl}=0$, gray). However, with ISS clipping (blue), the error remained within the theoretical bound and the amplification was suppressed under the same adversarial policy. Therefore, divergence of the system is structurally prevented even when the policy produces arbitrarily erroneous outputs.

Fig.~\ref{fig:iss}(b) shows that the state dependence of the bound behaved as intended, and that the choice of $\kappa$ was adequate in both directions. When a sinusoidal disturbance of gradually increasing amplitude was applied, the admissible range $\rho_k$ expanded as the state error increased and narrowed as the disturbance subsided, reflecting the dissipation margin available at each step. Throughout the run, $\rho_k$ remained above the margin required for residual compensation (red, $\lvert d_{\mathrm{res}}\rvert$); thus, legitimate compensation was never suppressed, whereas outside the saturated interval, it tracked $\lvert d_{\mathrm{res}}\rvert$ with a margin of the same order rather than opening an excessive gap. Under $\kappa=1$, the bound decreased to $1/3.5$ of the deployed value wherever the cap was inactive and was below $\lvert d_{\mathrm{res}}\rvert$ over most of the interval, confirming that the exact form~\eqref{eq:rho} was extremely conservative in allowing the required compensation. The saturation of $\rho_k$ at the design ceiling $\rho_{\max}$ during the peak interval was consistent with the design.

\begin{figure} 
    \centering
    \includegraphics[width=\linewidth]{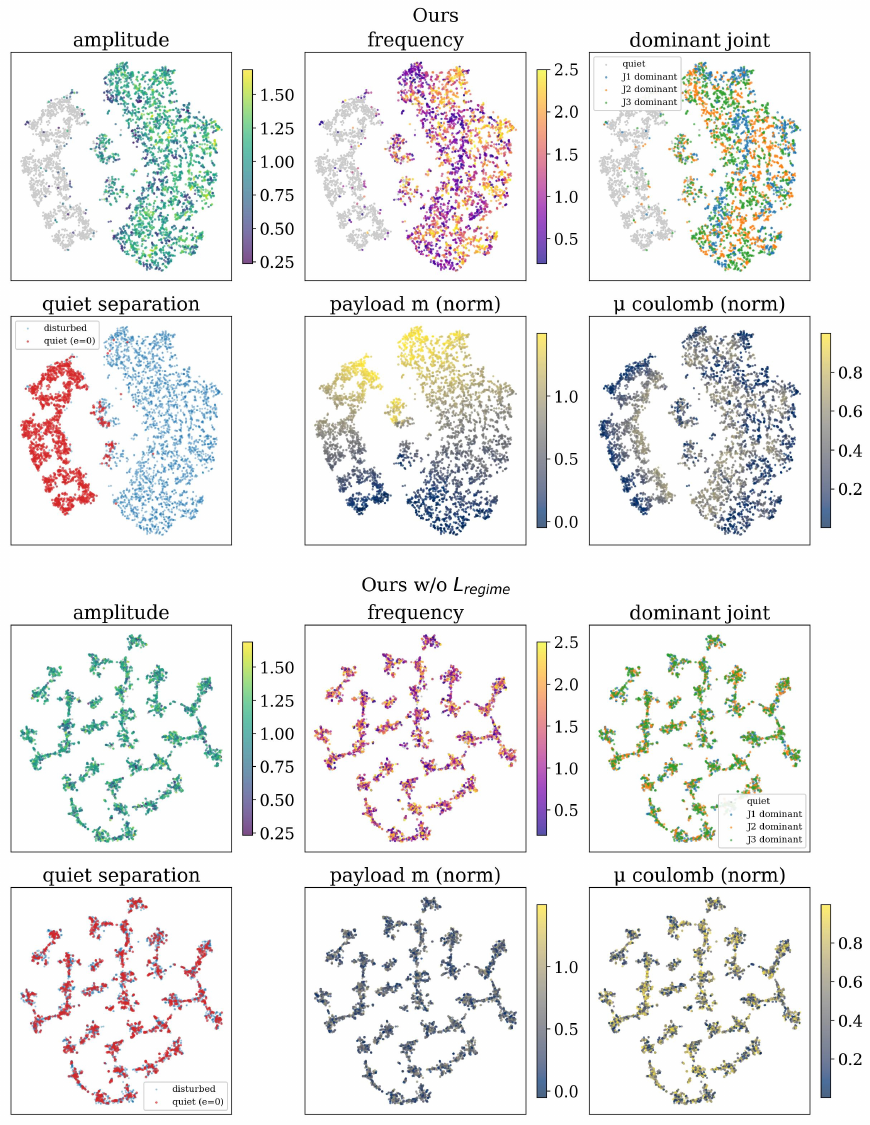} 
    \caption{t-SNE of the latent context $z$ colored by privileged context components. Top: proposed method. Bottom: Ours without $L_{regime}$.}
    \label{fig:latent}
\end{figure}

The ISS bound not only serves as a deployment time safeguard but also improves training itself. Fig.~\ref{fig:iss}(c) and (d) show comparisons of the three training curves, in which the components were cumulatively removed under identical conditions. As a common background, the estimation error of the model-based DOB (gray dashed line) increased over the iterations because the disturbance curriculum progressively increased the sinusoidal frequency range. The error after residual compensation (solid) decreased continuously; therefore, the gap from the DOB error directly represented the contribution of RL. Comparing Ours with Ours without ISS, the bound preemptively blocked catastrophic episodes generated by early exploration, thereby stabilizing the training. The configuration without the bound experienced a collapse, in which the reward decreased during training and required tens of iterations to recover; however, no such event was observed when the ISS bound was applied. Consequently, the final residual compensation error was lower than that without ISS, and the episode reward converged to a higher value with greater stability. This shows that the stability bound served not as a constraint that compromised performance, but as a filter that removed the extreme exploration negatively affecting the learning signal. Fig.~\ref{fig:iss}(c) and (d) include a configuration in which regime loss is also removed (Ours without ISS \& $\mathcal{L}_{\mathrm{regime}}$). This configuration exhibited the lowest performance in both the residual compensation error and episode reward, showing a further degradation on top of the gap caused by the removal of the ISS. The latent representation deprived of its alignment target led to this degradation, as confirmed by the latent-space analysis shown in Fig.~\ref{fig:latent}.

Fig.~\ref{fig:latent} shows the latent variable $z$ produced by the estimator projected onto two dimensions with t-SNE \cite{maaten2008} and the samples colored according to the components of the privileged context. In the proposed method, the latent space was organized by the disturbance regime, as shown at the top. First, the disturbance-free quiet condition (red) formed an independent region that was completely separated from the disturbed conditions, indicating that the disturbance-free reference point discussed in Section~\ref{sec:estimator} was realized as an explicit cluster. The payload mass formed a continuous gradient across the latent space, and the friction level likewise formed a banded structure according to its magnitude, confirming that in addition to discrete cluster separation, the magnitudes of the continuous parameters were encoded as the coordinates of the latent space. For this latent-space analysis, the friction was sampled uniformly over its normalized range $[0,1]$ instead of the Gaussian training distribution so that the friction channel spanned its full range. The color therefore denotes the sampled friction level of each episode. The amplitude and frequency of the sinusoidal disturbance also exhibited a value-dependent local structure within the disturbed region, together with clustering by the dominant joint.

Because the policy receives $z$ as a conditioning input, this spatial separation allows each region of the latent space to correspond to a distinct compensation strategy. The fast phase-matched compensation shown in Fig.~\ref{fig:dist_est}(a) indicates the direct result of this structure.

The bottom row shows the latent space trained without regime loss, namely, the previous approach \cite{himloco} that removed the context head and set the alignment target to the next observation $o_{t+1}$ instead of privileged labels. Although numerous fine-grained clusters appeared on the surface, the colors did not correspond to the clusters in any of the panels. The quiet samples were mixed with the disturbed samples across the entire space such that even the disturbance-free condition was not separated, and the payload and friction exhibited no spatial structure. Therefore, the clustering criterion was not a disturbance regime but regime-irrelevant factors, such as the trajectory phase. This result agrees exactly with the reason discussed in Section~\ref{sec:estimator}. The more effectively the NMPC and DOB cancel the disturbance, the more thoroughly their traces are erased from the next state. Therefore, under $o_{t+1}$ alignment, different disturbance conditions are superimposed onto the same phase orbit, and the latent representation loses the information needed to distinguish regimes. In such a latent space, the policy cannot condition its behavior on the disturbance type and is forced to retreat to an average compensation strategy over all regimes. This configuration exhibited the largest residual compensation error, as shown in Fig.~\ref{fig:iss}(c), illustrating the direct quantitative consequences of this representational collapse. The contrast between the two latent spaces demonstrates that the proposed regime loss is not merely auxiliary, but an essential component that enables regime-specialized compensation in an active disturbance-rejection architecture.

\begin{figure} 
    \centering
    \includegraphics[width=\linewidth]{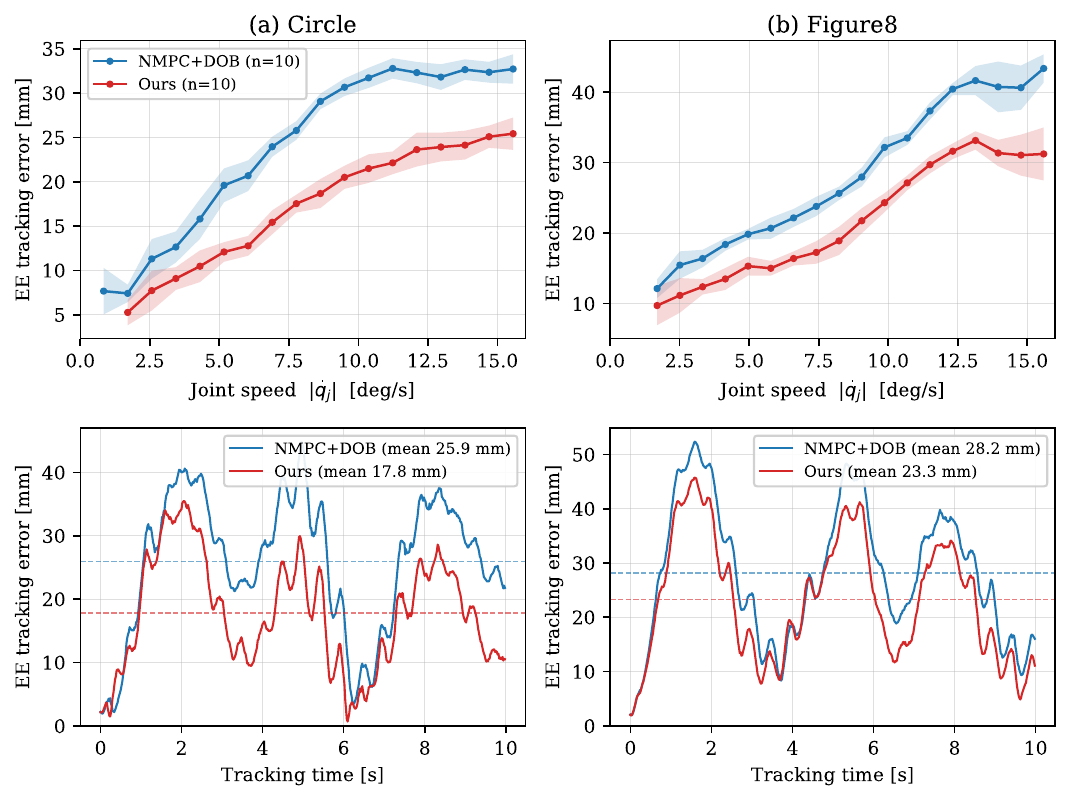} 
    \caption{Real-robot tracking under sinusoidal disturbance and a 500g countermass. Top: error versus joint speed, mean ± standard deviation of ten trials. Bottom: error time series of a representative trial.}
    \label{fig:real}
\end{figure}

\subsection{Real-Robot Experiments}
\label{sec:exp_real}
The improvement in trajectory tracking was confirmed using real hardware. We performed sim-to-real transfer to the PiPER manipulator and commanded circular and figure-eight trajectories under the simultaneous action of sinusoidal disturbances and a 500\,g countermass. The trajectory command was generated as $p^{*}(t) = \Gamma(\tau(t))$, where $\Gamma$ denotes the nominal path and the time warp defined by $\dot{\tau}(t) = 1 - 0.6\cos(2t) - 0.1\cos(3t)$ varied the traversal speed between approximately $0.3\times$ and $1.7\times$ the nominal speed without changing the path geometry, producing acceleration and deceleration conditions. Each trajectory experiment was repeated 10 times for each condition, and the top row in Fig.~\ref{fig:real} shows the tracking error averaged across 10 trials in the joint-speed bins (mean $\pm$ standard deviation), demonstrating repeatability.

The tracking error tended to increase with the joint speed. As the speed increased, the instantaneous effect of the sinusoidal disturbance on the trajectory increased, and unmodeled dynamic residuals, such as friction and inertial coupling, were amplified. A comparison across the entire growth range was important. In the top row of Fig.~\ref{fig:real}, the proposed method yields a consistently lower tracking error than the model-based DOB alone in every speed bin for both trajectories, and the gap between the two methods increases as the speed increases, which confirms that the contribution of the residual compensation increases precisely when the disturbance effect is the largest. Based on the average across ten trials, the tracking error decreased from $25.68 \pm 0.20$\,mm to $18.54 \pm 0.30$\,mm on the circular trajectory (27.8\% reduction) and from $27.77 \pm 0.22$\,mm to $23.12 \pm 0.31$\,mm on the figure-eight trajectory (16.7\% reduction).

The bottom row of Fig.~\ref{fig:real} shows the error time series of a representative trial, where the trial-mean error decreased from 25.9 to 17.8mm for the circle and from 28.2 to 23.3mm for the figure-eight. The reduction was largest in the acceleration and deceleration intervals, which is consistent with the speed-binned trend.


\begin{figure} 
    \centering
    \includegraphics[width=\linewidth]{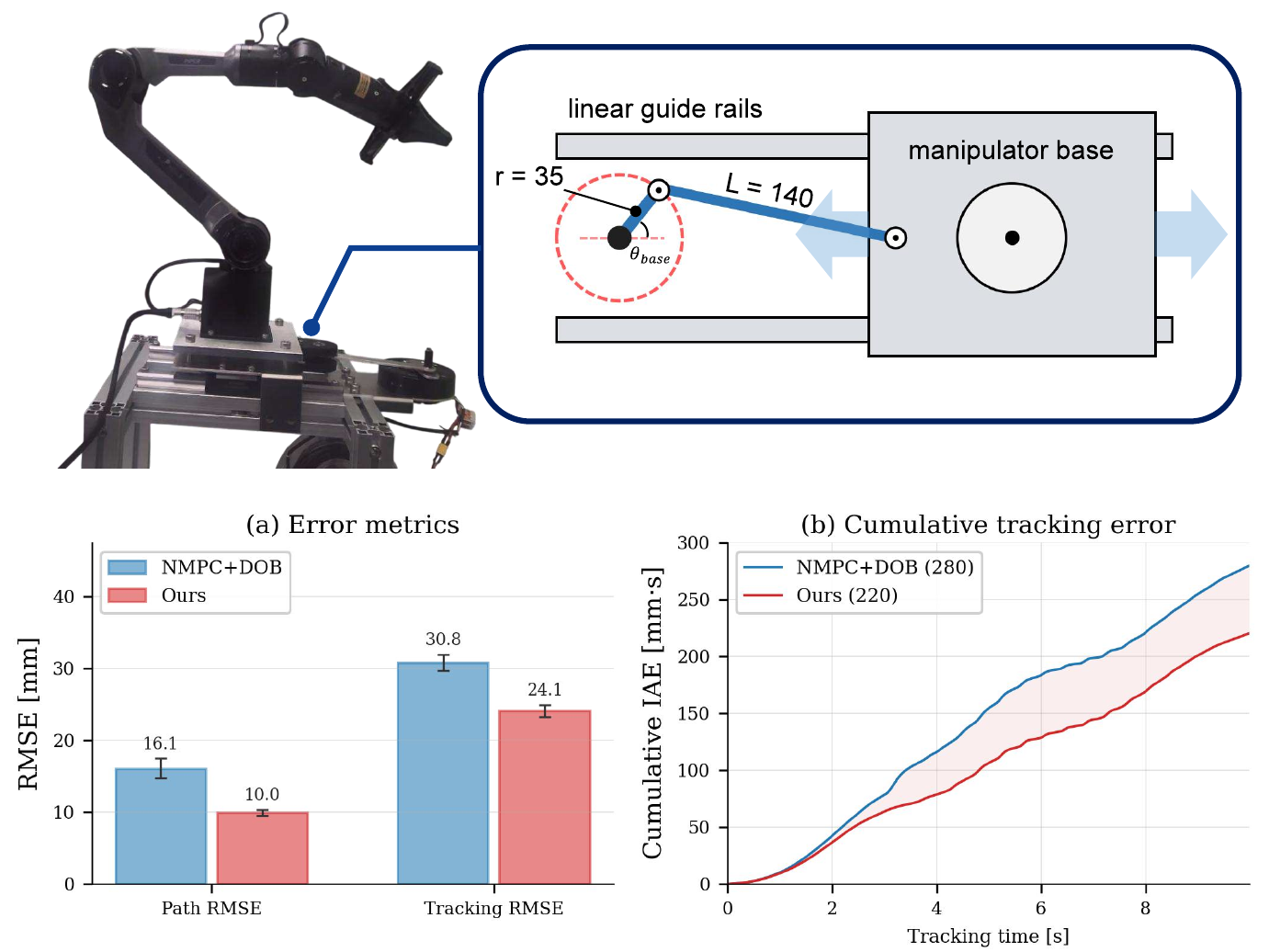} 
    \caption{Base-shaking experiment. Top: slider-crank bench that translates the manipulator base along linear rails. (a) path-geometry and tracking Root Mean Square Error(RMSE), and (b) cumulative Integral of Absolute Error(IAE).}
    \label{fig:base_shake}
\end{figure}

To emulate a disturbance that occurs during actual operation rather than an artificially injected joint torque, we excited the manipulator base with a vibrating motion. This disturbance source was not applied during training; therefore, the experiment examined how the learned policy and estimator responded to an unseen source. Base vibration was applied alone without any other disturbance; therefore, the observed performance difference was attributable entirely to the response to the new source.

As shown in Fig.~\ref{fig:base_shake} (top), the manipulator base was mounted on a slider running on linear guide rails and driven by translational vibration along the rail direction by an inline slider–crank mechanism consisting of a crank of radius $r_c$ and a coupler of length $L_c$. When the crank rotates at a constant angular velocity $\omega$, the rail-direction acceleration of the base at crank angle $\theta = \omega t$ is described as follows:
\begin{equation}
\ddot{x}_b(\theta) = -r_c\omega^{2}\left[\cos\theta + \frac{r_c}{L_c}\cdot\frac{\cos 2\theta + (r_c/L_c)^{2}\sin^{4}\theta} {\left(1 - (r_c/L_c)^{2}\sin^{2}\theta\right)^{3/2}}\right].
\label{eq:base_acc}
\end{equation}
In the experiment, a bench with $r_c = 35$\,mm and $L_c = 140$\,mm was driven at 150\,rpm, under which the base acceleration varied periodically between $-10.8$ and $+6.5$\,m/s$^2$. This acceleration appeared in the joint space as a disturbance torque. When the base translates without rotation with rail-direction acceleration $\ddot{x}_b$, the coupling terms caused by the base velocity cancel in the Lagrangian equations, and the fixed-base equation of motion \eqref{eq:dynamics} acquires only a single acceleration coupling term.
$\tau_{\mathrm{dist}} = \sum_i m_i J_{v,i}(q)^{\top}\mathbf{a}_b$, where
$\mathbf{a}_b$ is the base acceleration vector, whose rail direction
component is $\ddot{x}_b$, the remaining components are zero and
$m_i$ and $J_{v,i}$ are the mass and center-of-mass Jacobians of the link
$i$, respectively. The torque transmitted to joint $j$ is reduced to
\begin{equation}
\tau_{\mathrm{dist},j} = \bar{m}_j\,h_j(q)\,\ddot{x}_b,
\label{eq:base_torque}
\end{equation}
where $\bar{m}_j$ and $h_j(q)$ are the total mass and configuration-dependent moment arm of the links distal to joint $j$, respectively.
This relation shows that the base vibration is transmitted as a joint torque of the same form as the gravity term, with the gravity vector replaced by $-\mathbf{a}_b$, that is, a periodic inertial torque distributed to each joint with a magnitude that depends on the configuration. Consequently, although the base vibration is a new source, its signature in the joint space is a sinusoidal inertial torque whose dominant component lies within the frequency range of the sinusoidal disturbances applied during training. As the training environment was constructed to apply sinusoids, friction, payload, and impulses, this signature lies within the disturbance space represented by the training distribution.

Fig.~\ref{fig:base_shake} (bottom figure) shows the results of applying base vibration under a circular trajectory command. The proposed method reduced the path-geometry RMSE from 16.06\,mm to 9.96\,mm (38.0\% reduction) and the cumulative IAE from 280\,mm$\cdot$s to 220\,mm$\cdot$s (21.2\% reduction). The substantial improvement under these conditions can be explained by the composition of the disturbance. The dominant component of the inertial torque induced by the base vibration was above the passband of the DOB low-pass filter; therefore, the model-based DOB failed to estimate most of it, leaving it as a residual. In the proposed method, this residual was absorbed by the policy, and the outcome appeared as a reduction in both the path geometry error and cumulative error. This indicates that the division of roles in the proposed framework, in which the residual policy takes exclusive charge of the components that the DOB cannot handle, remains unchanged under an unseen disturbance source.

These results highlighted two important points. First, the compound disturbance environment constructed for training was adequate to represent the joint-space signature of disturbances arising in the actual operation because the trained policy was transferred to a new source without any retraining or parameter adjustment. Second, the estimator operated as intended for unseen sources. In Fig.~\ref{fig:base_shake}(b), the gap between the two curves forms within a few seconds of the onset of the tracking and enlarges monotonically thereafter. This indicates that the policy recognized the regime promptly and maintained compensation over the entire horizon, which reproduced on the hardware the rapid adaptation set as the design goal described in Section~\ref{sec:estimator}. For signatures outside the trained space, the ISS bound described in Section~\ref{sec:stability} still ensured a bounded tracking error regardless of regime misassignment. Therefore, the framework provides generalization and safety together.

\section{Conclusion}
\label{sec:conclusion}
This paper presents a residual reinforcement learning DOB framework in which the disturbance components that cannot be captured by a model-based DOB are assigned exclusively to an RL policy. An estimator network aligned with a privileged disturbance context organizes its latent space by disturbance regime, allowing the policy to recognize the acting disturbance and adapt rapidly across transitions. A state-dependent action bound derived from ISS analysis confines the tracking error to a certified envelope for arbitrary policy outputs through a closed-form scalar clip. Experiments on a 6-DOF manipulator confirmed the design at every level: the learned residual removed the phase lag and passband limitation of the DOB, the ISS bound suppressed adversarial amplification while stabilizing the training itself, and the policy transferred zero-shot to real hardware with a 27.8\% tracking error reduction, generalizing to a base-vibration disturbance not encountered during training with a 38.0\% reduction. The stability guarantee is conditional on the ISS properties of the baseline over its operating region, and the certified envelope reflects the relaxation introduced for practical compensation authority. Future studies will include tightening this envelope by replacing the worst-case alignment assumption with an estimate-informed bound, extending the validation to contact-rich tasks whose disturbance signatures lie outside the trained space, and comparing them with learning-based observer baselines on identical hardware.

\bibliographystyle{IEEEtran}
\bibliography{IEEEabrv,Bibliography}

@article{X1,
  title={Constrained model predictive control: Stability and optimality},
  author={Mayne, David Q and Rawlings, James B and Rao, Christopher V and Scokaert, Pierre OM},
  journal={Automatica},
  volume={36}, number={6}, pages={789--814}, year={2000},
  publisher={Elsevier}
}

@article{X2,
  title={Disturbance observer-based robust control and its applications: 35th anniversary overview},
  author={Sariyildiz, Emre and Oboe, Roberto and Ohnishi, Kouhei},
  journal={IEEE Transactions on Industrial Electronics},
  volume={67}, number={3}, pages={2042--2053}, year={2019},
  publisher={IEEE}
}

@misc{X4,
      title={Residual MPC: Blending Reinforcement Learning with GPU-Parallelized Model Predictive Control}, 
      author={Se Hwan Jeon and Ho Jae Lee and Seungwoo Hong and Sangbae Kim},
      year={2025},
      eprint={2510.12717},
      archivePrefix={arXiv},
      primaryClass={cs.RO},
      url={https://arxiv.org/abs/2510.12717}, 
}

@article{kamohara2025rl,
  title={RL-augmented adaptive model predictive control for bipedal locomotion over challenging terrain},
  author={Kamohara, Junnosuke and Wu, Feiyang and Wamorkar, Chinmayee and Hutchinson, Seth and Zhao, Ye},
  journal={arXiv preprint arXiv:2509.18466},
  year={2025}
}

@article{X5,
  title={Unsupervised learning of visual features by contrasting cluster assignments},
  author={Caron, Mathilde and Misra, Ishan and Mairal, Julien and Goyal, Priya and Bojanowski, Piotr and Joulin, Armand},
  journal={Advances in Neural Information Processing Systems},
  volume={33}, pages={9912--9924}, year={2020}
}

@inproceedings{X7,
  title={High-frequency nonlinear model predictive control of a manipulator},
  author={Kleff, S{\'e}bastien and Meduri, Avadesh and Budhiraja, Rohan and Mansard, Nicolas and Righetti, Ludovic},
  booktitle={2021 IEEE International Conference on Robotics and Automation (ICRA)},
  pages={7330--7336},
  year={2021},
  organization={IEEE}
}

@article{X8,
  title={Motion control for advanced mechatronics},
  author={Ohnishi, Kouhei and Shibata, Masaaki and Murakami, Toshiyuki},
  journal={IEEE/ASME transactions on mechatronics},
  volume={1},
  number={1},
  pages={56--67},
  year={1996},
  publisher={IEEE}
}

@inproceedings{X9,
  title={A robust decentralized joint control based on interference estimation},
  author={Nakao, Masato and Ohnishi, Kouhei and Miyachi, Kunio},
  booktitle={Proceedings 1987 IEEE International Conference on Robotics and Automation},
  volume={4}, pages={326--331}, year={1987}, organization={IEEE}
}

@article{X10,
  title={A nonlinear disturbance observer for robotic manipulators},
  author={Chen, Wen-Hua and Ballance, Donald J and Gawthrop, Peter J and O'Reilly, John},
  journal={IEEE Transactions on Industrial Electronics},
  volume={47}, number={4}, pages={932--938}, year={2000},
  publisher={IEEE}
}

@article{X11,
  title={Nonlinear disturbance observer design for robotic manipulators},
  author={Mohammadi, Alireza and Tavakoli, Mahdi and Marquez, Horacio J and Hashemzadeh, Farzad},
  journal={Control Engineering Practice},
  volume={21}, number={3}, pages={253--267}, year={2013},
  publisher={Elsevier}
}

@article{X12,
  title={Design of a momentum-based disturbance observer for rigid and flexible joint robots},
  author={Kim, Min Jun and Park, Young Jin and Chung, Wan Kyun},
  journal={Intelligent Service Robotics},
  volume={8},
  number={1},
  pages={57--65},
  year={2015},
  publisher={Springer}
}

@inproceedings{X15,
  title={Deep reinforcement learning for robotic manipulation with asynchronous off-policy updates},
  author={Gu, Shixiang and Holly, Ethan and Lillicrap, Timothy and Levine, Sergey},
  booktitle={2017 IEEE International Conference on Robotics and Automation (ICRA)},
  pages={3389--3396}, year={2017}, organization={IEEE}
}

@misc{x16,
      title={IndustReal: Transferring Contact-Rich Assembly Tasks from Simulation to Reality}, 
      author={Bingjie Tang and Michael A. Lin and Iretiayo Akinola and Ankur Handa and Gaurav S. Sukhatme and Fabio Ramos and Dieter Fox and Yashraj Narang},
      year={2023},
      eprint={2305.17110},
      archivePrefix={arXiv},
      primaryClass={cs.RO},
      url={https://arxiv.org/abs/2305.17110}, 
}

@article{X17,
  title={Learning force control for contact-rich manipulation tasks with rigid position-controlled robots},
  author={Beltran-Hernandez, Cristian Camilo and Petit, Damien and Ramirez-Alpizar, Ixchel Georgina and Nishi, Takayuki and Kikuchi, Shinichi and Matsubara, Takamitsu and Harada, Kensuke},
  journal={IEEE Robotics and Automation Letters},
  volume={5}, number={4}, pages={5709--5716}, year={2020},
  publisher={IEEE}
}

@inproceedings{X19,
  title={DOB-Net: Actively rejecting unknown excessive time-varying disturbances},
  author={Wang, Tianming and Lu, Wenjie and Yan, Zheng and Liu, Dikai},
  booktitle={2020 IEEE International Conference on Robotics and Automation (ICRA)},
  pages={1881--1887}, year={2020}, organization={IEEE}
}

@article{hewing2020learning,
  title={Learning-based model predictive control: Toward safe learning in control},
  author={Hewing, Lukas and Wabersich, Kim P and Menner, Marcel and Zeilinger, Melanie N},
  journal={Annual Review of Control, Robotics, and Autonomous Systems},
  volume={3},
  number={1},
  pages={269--296},
  year={2020},
  publisher={Annual Reviews}
}

@article{X20,
  title={DiAReL: Reinforcement learning with disturbance awareness for robust sim2real policy transfer in robot control},
  author={Malmir, Mohammadhossein and Josifovski, Josip and Klarmann, Noah and Knoll, Alois},
  journal={IEEE Transactions on Control Systems Technology},
  year={2025}, publisher={IEEE}
}

@article{X21,
  title={EVOLVER: Online learning and prediction of disturbances for robot control},
  author={Jia, Jindou and Zhang, Wenyu and Guo, Kexin and Wang, Jianliang and Yu, Xiang and Shi, Yang and Guo, Lei},
  journal={IEEE Transactions on Robotics},
  volume={40},
  pages={382--402},
  year={2023},
  publisher={IEEE}
}

@misc{X22,
      title={Residual Policy Learning for Vehicle Control of Autonomous Racing Cars}, 
      author={Raphael Trumpp and Denis Hoornaert and Marco Caccamo},
      year={2023},
      eprint={2302.07035},
      archivePrefix={arXiv},
      primaryClass={cs.RO},
      url={https://arxiv.org/abs/2302.07035}, 
}

@article{X23,
  title={Data-driven model predictive control for trajectory tracking with a robotic arm},
  author={Carron, Andrea and Arcari, Elena and Wermelinger, Martin and Hewing, Lukas and Hutter, Marco and Zeilinger, Melanie N},
  journal={IEEE Robotics and Automation Letters},
  volume={4}, number={4}, pages={3758--3765}, year={2019},
  publisher={IEEE}
}

@inproceedings{X24,
  title={Actor-critic model predictive control},
  author={Romero, Angel and Song, Yunlong and Scaramuzza, Davide},
  booktitle={2024 IEEE International Conference on Robotics and Automation (ICRA)},
  pages={14777--14784}, year={2024}, organization={IEEE}
}

@inproceedings{cuturi,
  title={Sinkhorn distances: Lightspeed computation of optimal transport},
  author={Cuturi, Marco},
  booktitle={Advances in Neural Information Processing Systems},
  volume={26}, year={2013}
}

@inproceedings{kendall2018,
  title={Multi-task learning using uncertainty to weigh losses for scene geometry and semantics},
  author={Kendall, Alex and Gal, Yarin and Cipolla, Roberto},
  booktitle={Proceedings of the IEEE Conference on Computer Vision and Pattern Recognition},
  pages={7482--7491}, year={2018}
}

@inproceedings{himloco,
  title={Hybrid Internal Model: Learning Agile Legged Locomotion with Simulated Robot Response},
  author={Long, Junfeng and Wang, ZiRui and Li, Quanyi and Cao, Liu and Gao, Jiawei and Pang, Jiangmiao},
  booktitle={The Twelfth International Conference on Learning Representations},
  year={2024}
}

@incollection{sontag2008,
  title={Input to state stability: Basic concepts and results},
  author={Sontag, Eduardo D},
  booktitle={Nonlinear and Optimal Control Theory},
  pages={163--220}, year={2008}, publisher={Springer}
}

@article{maaten2008,          
  title={Visualizing data using t-SNE},
  author={van der Maaten, Laurens and Hinton, Geoffrey},
  journal={Journal of Machine Learning Research},
  volume={9}, pages={2579--2605}, year={2008}
}

@book{rubinstein2004,
  title={The cross-entropy method: a unified approach to combinatorial optimization, Monte-Carlo simulation, and machine learning},
  author={Rubinstein, Reuven Y and Kroese, Dirk P},
  volume={133},
  year={2004},
  publisher={Springer}
}

@article{wabersich2021,
  title={A predictive safety filter for learning-based control of constrained nonlinear dynamical systems},
  author={Wabersich, Kim P and Zeilinger, Melanie N},
  journal={Automatica}, volume={129}, pages={109597}, year={2021}
}

@inproceedings{johannink2019residual,
  title={Residual reinforcement learning for robot control},
  author={Johannink, Tobias and Bahl, Shikhar and Nair, Ashvin and Luo, Jianlan and Kumar, Avinash and Loskyll, Matthias and Ojea, Juan Aparicio and Solowjow, Eugen and Levine, Sergey},
  booktitle={2019 international conference on robotics and automation (ICRA)},
  pages={6023--6029},
  year={2019},
  organization={IEEE}
}

@inproceedings{tobin2017,
  title={Domain randomization for transferring deep neural networks from simulation to the real world},
  author={Tobin, Josh and Fong, Rachel and Ray, Alex and Schneider, Jonas and Zaremba, Wojciech and Abbeel, Pieter},
  booktitle={2017 IEEE/RSJ International Conference on Intelligent Robots and Systems (IROS)},
  pages={23--30}, year={2017}, organization={IEEE}
}

@article{silver2018,          
  title={Residual policy learning},
  author={Silver, Tom and Allen, Kelsey and Tenenbaum, Josh and Kaelbling, Leslie},
  journal={arXiv preprint arXiv:1812.06298}, year={2018}
}

@article{nikoobin2009,
  title={Lyapunov-based nonlinear disturbance observer for serial n-link robot manipulators},
  author={Nikoobin, Amin and Haghighi, Reza},
  journal={Journal of Intelligent and Robotic Systems},
  volume={55}, number={2}, pages={135--153}, year={2009}
}

@article{bjelonic2025towards,
   title={Towards bridging the gap: Systematic sim-to-real transfer for diverse legged robots},
   ISSN={1741-3176},
   url={http://dx.doi.org/10.1177/02783649261459628},
   DOI={10.1177/02783649261459628},
   journal={The International Journal of Robotics Research},
   publisher={SAGE Publications},
   author={Bjelonic, Filip and Tischhauser, Fabian and Hutter, Marco},
   year={2026},
   month=July }

\end{document}